\documentclass{article}

\usepackage{PRIMEarxiv}

\usepackage[utf8]{inputenc} % allow utf-8 input
\usepackage[T1]{fontenc}    % use 8-bit T1 fonts
\usepackage{hyperref}       % hyperlinks
\usepackage{url}            % simple URL typesetting
\usepackage{booktabs}       % professional-quality tables
\usepackage{amsfonts}       % blackboard math symbols
\usepackage{nicefrac}       % compact symbols for 1/2, etc.
\usepackage{microtype}      % microtypography
\usepackage{lipsum}
\usepackage{fancyhdr}       % header
\usepackage{graphicx}       % graphics
\graphicspath{{media/}}     % organize your images and other figures under media/ folder
\usepackage{geometry}
\usepackage{amsmath,amsfonts}
\usepackage{amssymb}
\usepackage{bm}
\usepackage{lipsum}
\usepackage{algorithm}
\usepackage{algpseudocode}
\usepackage{comment}
\usepackage{subcaption}
\usepackage{cite}
\title{Discovering governing equation in  structural dynamics from acceleration-only measurements
}

\author{
  Calvin Alvares \\
  Department of Applied Mechanics \\
  Indian Institute of Technology Delhi \\
  Hauz Khas - 110016, India\\
  \texttt{alvarescalvin16@gmail.com} \\
   \And
  Souvik Chakraborty \\
  Department of Applied Mechanics\\ Yardi School of Artificial Intelligence \\
  Indian Institute of Technology Delhi \\
  Hauz Khas - 110016, India\\
  \texttt{souvik@am.iitd.ac.in} \\
}

\begin{document}
\maketitle

\begin{abstract}
Over the past few years, equation discovery has gained popularity in different fields of science and engineering. However, existing equation discovery algorithms rely on the availability of noisy measurements of the state variables (i.e., displacement {and velocity}). This is a major bottleneck in structural dynamics, where we often only have access to acceleration measurements. To that end, this paper introduces a novel equation discovery algorithm for discovering governing equations of dynamical systems from acceleration-only measurements. The proposed algorithm employs a library-based approach for equation discovery. To enable equation discovery from acceleration-only measurements, we propose a novel Approximate Bayesian Computation (ABC) model that prioritizes parsimonious models. The efficacy of the proposed algorithm is illustrated using {four} structural dynamics examples that include both linear and nonlinear dynamical systems. The case studies presented illustrate the possible application of the proposed approach for equation discovery of dynamical systems from acceleration-only measurements.
\end{abstract}

% keywords can be removed
\keywords{equation discovery \and approximate Bayesian computation \and nonlinear system identification \and model selection}

\section{Introduction}
\label{sec:intro}
Traditionally, dynamical system models are discovered from first principles and prior knowledge of the nature of a system, and, such models have been very successfully used for the design, analysis and prediction of complex dynamical systems. However, for many systems, such as the human brain, power grids, climate systems, and ecosystems, identifying a dynamical system model solely from first principles may be difficult. Fortunately, advancements in machine learning, alongside an increase in computing power and data collection techniques, have fueled the adoption of a new paradigm in the discovery of dynamical system models -- one driven by data %\cite{bongard2007automated, schmidt2009distilling, brunton2016discovering, mangan2016inferring, schaeffer2017sparse, mangan2017model, kaiser2018sparse, champion2019data, schaeffer2020extracting, boninsegna2018sparse, mangan2019model, stender2019recovery, rudy2017data, zhang2018robust, chen2020deep, rudy2019deep, raissi2018multistep, raissi2018deep, nayek2021spike}
\cite{brunton2016discovering}%, mangan2017model, schaeffer2020extracting, stender2019recovery, chen2020deep}
. For a dynamical system represented by $\dot{x}(t) = F(x(t))$, where $x$ is the state of the system, and $F(x(t))$ is a function of the state that encodes the system's dynamics, data-driven equation discovery seeks to uncover the form of $F(x(t))$ from time-series data of the system. This involves two simultaneous tasks -- model identification, which involves finding a suitable form for $F(x(t))$, and parameter estimation, which involves determining the parameters of the chosen model. A critical requirement enforced on data-driven dynamical system discovery methods is that they should discover parsimonious models by finding the least complex model which sufficiently fits the measured data. This ensures that the discovered model is free from overfitting and, thus, has good predictive power. Furthermore, such models are interpretable by virtue of having a limited number of features, thus providing a useful understanding of the underlying system dynamics. 

Early attempts at data-driven equation discovery used symbolic regression \cite{bongard2007automated, schmidt2009distilling} to identify suitable functions from a library of candidate functions. However, symbolic regression is computationally expensive, does not scale well to large problems and is prone to overfitting. Following this, \textit{sparse identification of nonlinear dynamics (SINDy)} \cite{brunton2016discovering} was developed. It relies on having a library of candidate functions and using sparse regression to select a few important terms from the library that accurately reflect the system's dynamics. This approach is scalable and leads to the rapid discovery of parsimonious and interpretable models. SINDy has been widely adopted for data-driven dynamical system discovery and has been further developed in various studies. It has been used for identifying ordinary differential equations (ODEs) \cite{brunton2016discovering} and partial differential equations (PDEs) \cite{rudy2017data, schaeffer2017learning}; modelling of fluid dynamics \cite{loiseau2018constrained, loiseau2018sparse}, biological networks 
\cite{mangan2016inferring}, plasma dynamics \cite{dam2017sparse}, nonlinear optics \cite{sorokina2016sparse}, chemical dynamics \cite{narasingam2018data}, structural dynamics \cite{lai2019sparse}; sparse identification with control \cite{kaiser2018sparse}, sparse identification with rational functions \cite{kaheman2020sindy},   model selection using an integral
formulation of differential equations \cite{schaeffer2017sparse}, discovery of coordinates for sparse representation of governing equations \cite{champion2019data}, {sparse learning of stochastic dynamic equations \cite{boninsegna2018sparse, tripura2023bayesian}, model selection for hybrid dynamical systems \cite{mangan2019model},} amongst others. Apart from SINDy, black-box approaches using deep neural networks \cite{rudy2019deep, raissi2018multistep, raissi2018deep} have been used for nonlinear system identification. However, these approaches are used for prediction and do not explicitly provide the governing equations. Bayesian approaches have also been used for the discovery of governing equations \cite{zhang2018robust, nayek2021spike, fuentes2021equation, more2023bayesian, tripura2023sparse, north2022bayesian} by utilizing their ability to perform simultaneous model selection and parameter estimation. An advantage of using a Bayesian framework is that it can capture the uncertainty in the estimated parameters and easily eliminate the possibility of overfitting the data by the use of sparsity-promoting priors. Application of Bayesian equation discovery can be found in reliability analysis \cite{mathpati2023mantra}, discovery of continuous dynamical systems \cite{more2023bayesian}, discovery of stochastic dynamical systems \cite{tripura2023sparse}, discovery of lagrangian of dynamical 
and stochastic systems \cite{tripura2023bayesian} and discovery of stochastic partial differential equations \cite{mathpati2024discovering}, among others.

All established approaches for data-driven equation discovery rely on the availability of displacement {and velocity} measurements. However, 
in dynamical systems, we often have access to acceleration data measured using accelerometers. Therefore, from a practical standpoint, a realistic setup would be to discover the governing equations of dynamical systems using only acceleration measurements.
%For equation discovery using the time derivative data of the state variables, the use of any existing equation discovery method is unfeasible as obtaining the state variables with reasonable accuracy from the time derivative data of the state variables using numerical integration is not possible due to measurement data being corrupted with noise.
%
In response, we here propose an  \textit{Approximate Bayesian Computation} (ABC)  \cite{marin2012approximate} based equation discovery approach that discovers underlying equations solely using the acceleration measurements. With ABC, we eliminate the need for constructing the likelihood function that in turn is dependent on the state variables (displacement {and velocity}). We propose a variant of ABC that exploits spike and slab prior \cite{mitchell1988bayesian} to learn a parsimonious equation of dynamical systems from acceleration measurements. Here onwards, we refer to the proposed ABC algorithm as sparse ABC (s-ABC). We illustrate the efficacy of s-ABC using a number of benchmark dynamical systems from the literature. 

% Likelihood computation in Bayesian parameter estimation requires information of the state \cite{nayek2021spike, fuentes2021equation}; however, ABC  does not require the evaluation of the likelihood of the data and, instead, relies on the similarity of simulated data from generated models to the observed data. Based on this similarity, the correct model and its parameters can be estimated. Thus, eliminating likelihood computation in ABC makes it possible for it to discover the governing equations of a system solely from time derivative data. Through the use of a novel ABC approach known as \textit{sparse approximate Bayesian computation (S-ABC)}, we show that the proposed approach successfully discoverers the governing equations of various systems from time derivative data.

The remainder of the paper is organized as follows. The problem formulation is described in Section \ref{sec:sec2}. Section \ref{sec:sec3} describes the proposed s-ABC approach. Section \ref{sec:sec4} illustrates the performance of the proposed approach in discovering governing equations from acceleration-only measurements. Finally, Section \ref{sec:sec5} discusses and concludes the work.

\section{Problem Formulation}\label{sec:sec2}
We represent a generic dynamical system as:
\begin{equation}\label{eq:dyn}
    \dot {\bm {x}}(t) = F\left(\bm {x}(t)\right),
\end{equation}
where $\bm x(t) = [x_1(t), x_2(t), ..., x_n(t)]^T \in \mathbb{R}^{n \times 1}$ is the state of the system at time $t$ and $F\left(\bm x(t)\right)$ is a function that governs the dynamic evolution of the system. 
Without loss of generality, we consider a single-degree-of-freedom dynamical system,
\begin{equation}\label{eq:sdof}
    m\ddot u + c \dot u + k u + g(u, \dot u) = f,
\end{equation}
where $m$, $c$, and $k$ represent the mass, damping and stiffness, $g(u, \dot u)$ is a generic nonlinear function of displacement $u$ and velocity $\dot u$, and is responsible for any nonlinear behavior present in the single-degree-of-freedom system in Eq. \eqref{eq:sdof}. $f$ is Eq. \eqref{eq:sdof}  represents the forcing function and $\ddot u $ represents the acceleration. Note that Eq. \eqref{eq:sdof} can be recast into the generic form shown in Eq. \eqref{eq:dyn} by considering
\begin{subequations}\label{eq:trans_var}
    \begin{equation}
        x_1 = u, 
    \end{equation}
    \begin{equation}
        x_2 = \dot u.
    \end{equation}
\end{subequations}
With this setup, Eq. \eqref{eq:sdof} takes the following form
\begin{subequations}\label{eq:ssm}
    \begin{equation}\label{eq:ssm1}
        \dot x_1 = x_2,
    \end{equation}
    \begin{equation}\label{eq:ssm2}
        \dot x_2 = \frac{1}{m}\left(f - c x_2 - k x_1 - g(x_1, x_2) \right).
    \end{equation}
\end{subequations}
Note that Eq. \eqref{eq:ssm1} is just another representation of velocity and hence, can be ignored. Therefore, the discovery of the governing equation from data is equivalent to discovering the right-hand side of Eq. \eqref{eq:ssm2}. 
%
%
% of the system, respectively. $u$, $\dot u$, $\ddot u$ represents the acceleration, $\dot u$ represents the velocity, and $u$ represents the displacement. 
%
%
%
% We seek to identify a dynamical systems given by 
% \begin{equation}
%     \dot{x}(t) = F(x(t))
% \end{equation}
% where $x(t) = [x_1(t), x_2(t), ..., x_n(t)]^T \in \mathbb{R}^{n \times 1}$ is the state of the system at time $t$, and $F(x(t))$ is an unknown function of the state $x(t)$. Prior information about the system may result in the knowledge of one or more equations of the dynamical system, and thus, the entirety of $F(x(t))$ may not be unknown.
%
% To that end, the basic idea behind library-based approach for equation discovery is to 
%
%
% To uncover the full form of $F(x(t))$, we 
consider a dictionary of $p$ candidate functions given by 
\begin{equation}
\bm B = [f_1(x_1,x_2), f_2(x_1,x_2), ..., f_p(x_1,x_2)].
\end{equation} 
Using these potential candidate functions, the unknown Eq. \eqref{eq:ssm2} is represented as
\begin{equation}\label{eq:edps}
    \dot{x_2} \approx \theta_{1} f_1(x_1,x_2)+\theta_{2} f_2(x_1,x_2)+\ldots+\theta_{p} f_p(x_1,x_2).
\end{equation}
In the above expression, $\bm \theta = \left\{\theta_{i}\right\}_{i=1}^p$ represents the weight vector.  For most systems, $\bm \theta$ is a sparse vector with the non-zero terms indicative of system parameters. With this setup, equation discovery reduces to determining (a) which $\theta_i$ are active (the model selection problem) and (b) the value of the active $\theta_i$ (the parameter identification problem).

Existing methods for solving Eq. \eqref{eq:edps} rely on noisy measurements of the displacement $u = x_1$. However, often in structural dynamics, we only have access to the acceleration measurements $\dot x_2 = \ddot u$. The objective here is to develop an algorithm for finding the vector $\bm \theta$ in Eq. \eqref{eq:edps} from acceleration-only measurements.

\section{Sparse approximate Bayesian computation } \label{sec:sec3}
In response to the objective specified in Section \ref{sec:sec2}, we propose a novel Approximate Bayesian computation \cite{csillery2010approximate} based equation discovery framework. Conventional Bayesian parameter estimation involves estimating parameters based on prior beliefs about the parameters and observed data, and for the task of equation discovery, this has been successfully used \cite{nayek2021spike, fuentes2021equation, more2023bayesian, tripura2023sparse, north2022bayesian}. For Bayesian parameter estimation, the posterior distribution of parameters $\bm \theta$, conditioned on the observed data $\bm D $, can be expressed as
\begin{equation}
    P(\bm \theta|\bm D) = \frac{P(\bm D|\bm \theta)P(\bm \theta)}{P(\bm D)},
\end{equation}
where $P(\bm D|\bm \theta)$ is the likelihood of the data, $P(\bm \theta)$ is the parameter prior distribution, and $P(\bm D)$ is the data evidence. For the parameter posterior estimation, $P(\bm D)$ is intractable and hence, is bypassed by using either sampling-based \cite{nayek2021spike} or variational \cite{more2023bayesian} approaches. It is, however, necessary to evaluate the likelihood $P(\bm D|\bm \theta)$. Unfortunately, for the problem setup considered in this paper, the likelihood $P(\bm D|\bm \theta)$ is intractable, and hence, conventional Bayesian approaches cannot be employed. Approximate Bayesian computation (ABC) addresses this issue by estimating the parameter posterior without explicitly computing the likelihood \cite{marin2012approximate} and has been successfully used for problems where the likelihood is computationally intractable or difficult to compute \cite{marin2012approximate, beaumont2010approximate, beaumont2002approximate, akeret2015approximate, drovandi2011estimation, technow2015integrating}.
% Since we are provided with noisy data of the time derivative of the state variables, obtaining the state variables through numerical integration is not feasible. Since the data of the state variables are unavailable, the likelihood $P(D|\theta)$ is intractable, and thus computing the parameter posterior is not possible. Approximate Bayesian computation (ABC) addresses this issue by estimating the parameter posterior without explicitly computing the likelihood \cite{marin2012approximate} and has been successfully used for problems where the likelihood is computationally intractable or too difficult to compute \cite{marin2012approximate, beaumont2010approximate, beaumont2002approximate, akeret2015approximate, drovandi2011estimation, technow2015integrating}. 
On a fundamental level, ABC involves generating simulated data for different parameter values and comparing the simulated data $\bm D^*$ to the observed data $\bm D$ using a distance measure $\rho(\bm D, \bm D^*)$. Parameters are accepted if the distance of the simulated data to the observed data is within a user-defined tolerance $\epsilon$. Using ABC, the parameter posterior can be estimated as
\begin{equation}
    p (\bm \theta|\bm D) \approx p_{\epsilon} (\bm \theta|\bm D) \propto \int p(\bm \theta, \bm D^*|\rho(\bm D,\bm D^*)< \epsilon) \: d \bm D^*,
\end{equation}
where $p_{\epsilon} (\bm \theta|\bm D)$ is the ABC-based parameter posterior which tends to the true parameter posterior as $\epsilon$ tends to zero. 

Application of ABC can also be found for simultaneous model selection and parameter estimation \cite{liepe2014framework, abdessalem2018model, toni2009approximate, abdessalem2019model, nayek2023identification}. This is achieved by using ABC to select a model and its parameters from a small set of predefined models based on the acceptance frequency of parameters for the different models. The number of possible models for a system with $p$ basis functions is $2^{p}$. Thus, for a sufficiently large dictionary, the task of equation discovery amounts to selecting a model from an extremely large set of possible models, along with simultaneous parameter estimation. Popular ABC algorithms existing in the literature include ABC rejection sampling \cite{pritchard1999population}, ABC Sequential Monte Carlo \cite{toni2009approximate}, ABC-MCMC \cite{marjoram2003markov}, ABC Nested Sampling \cite{abdessalem2019model} and ABC Subset Simulation \cite{chiachio2014approximate}. However, these algorithms are often not scalable. Here, we propose a novel sparse ABC (s-ABC) algorithm for equation discovery using acceleration-only measurements. The fundamental idea in s-ABC is to conduct parameter estimation while invoking the principle of parsimony to ensure that the predicted parameter vector is sufficiently sparse. In s-ABC, we use a nested sampling approach along with a number of sparsity-promoting measures, including sampling from sparsity-promoting prior, regularization, and reinitialization to encourage exploration. 
% Overall, the proposed s-ABC involves generation of initial population and updating of the particles. We introduce targeted modification in both the steps to encourage sparsity.
\subsection{Initial population generation}\label{subsec:proposal}
% to identify the correct model from the large set of models that are expressed by the different combinations of the basis functions in the dictionary, we conduct parameter estimation while invoking the principle of parsimony to ensure that predicted parameter vector is sufficiently sparse. This is done by our proposed algorithm --- sparse approximate Bayesian computation (S-ABC), in which we use a nested sampling approach \cite{nayek2023identification} along with a number of sparsity-promoting measures. The next few paragraphs outline this algorithm.
%
The first stage involves generating an initial population of $N_S$ independent sets of parameters ($\bm \theta \in \mathbb{R}^{N}$). Here onwards, we refer to an independent set of parameters as a particle and the collection of all particles as a population. We propose a three-step sparsity-promoting strategy to ensure sparsity in the initial population. In the first step, we generate the initial population from a sparsity promoting prior. Although a number of sparsity-promoting priors exist in the literature, we here employ a variant of the spike-and-slab prior for generating the initial population
\begin{equation}\label{eq:ss}
    P\left( \bm \theta | \bm Z \right) = P_{\text{slab}}(\bm \theta _k)\prod_{i,Z_i=0} P_{\text{spike}}\left( \theta _i\right),        
\end{equation}
where
\begin{equation}\label{eq:spike}
    P_{\text{spike}}\left( \theta _i\right) = \delta_0.
\end{equation}
With this setup, the parameters that belong to the spike are assigned zero values. $Z_i$ in Eq. \eqref{eq:ss} is a binary-valued random variable that follows Bernoulli distribution
\begin{equation}\label{eq:bern}
    P(Z_i|\eta) = \text{Bern}(\eta),\; \; i = 1,\ldots, N.
\end{equation}
This ensures that a parameter will belong to the spike with probability $\eta$. The proposed approach is highly flexible and allows the user to decide the slab distribution. While in theory, any valid distribution can be used as $P_{\text{slab}}\left(\bm \theta_k\right)$, we have used uniform distribution,  $P_{\text{slab}}\left(\bm \theta_k\right)$ for initial population generation.

In the second step, we set non-zero parameters with small magnitudes as zero. This is ensured by introducing a hyperparameter $\lambda$ such that
\begin{equation}\label{eq:sparse2}
    \theta_i = \left\{ \begin{array}{cl}
        0 & \text{if } |\theta_i| \le \lambda \\
         \theta_i & \text{elsewhere}
    \end{array} \right.
\end{equation}
Note that for enhanced flexibility, it is possible to use different thresholds for different particles. 
% \blue{To do that we have defined the hyperparameter vector $\bm \lambda = [\lambda_1, \lambda_2, ..., \lambda_N]$, where $\lambda_i$ is the the thresholding hyperparameter corresponding to the parameter $\theta_i$}

In the last step, each of the particles generated, $\theta^* _i$ are substituted into the RHS of Eq. \eqref{eq:edps} and simulated to obtain the simulated response $\bm D^*$. The best particles from the population are selected based on a distance metric. A commonly used distance function is the normalised mean square error (NMSE). However, to encourage sparsity, we define the distance function (or loss) by adding an additional sparsity-promoting term to the NMSE,
% \begin{equation}\label{eq:rNMSE}
%     \rho(D,D^*) = \rho(D,D^*,\beta, \theta^*) = \sum_{j \in S} \frac{100}{m \sigma_j^2}\sum_{i=1}^{m} (\dot{x}^*_{j}(t_i)-\dot{x}_{j}(t_i))^2 + \beta ||\theta^*||_0
% \end{equation}
% \blue
\begin{equation}\label{eq:rNMSE}
    \rho(\bm D,\bm D^*) = \rho(\bm D,\bm D^*,\beta, \bm \theta^*) = \frac{100}{m \sigma_{\bm D}^2}\sum_{i=1}^{m} (\dot{x}^*_{2}(t_i)-\dot{x}_{2}(t_i))^2 + \beta ||\bm \theta^*||_0,
\end{equation}
%
% To generate every particle, parameters $\bm \theta^*$ are chosen from predefined parameter priors, $p\left(\bm \theta\right)$. Then, with a probability $\eta$, each parameter ($ \theta^*_i$) is independently set to zero in order to promote sparsity. $\lambda \in \mathbb{R}^{N}$ is a sparsity-promoting hyperparameter vector with the same length as the parameter vector. Parameter values $\theta^*[i]$ less than $\lambda[i]$ are set to zero. After this, simulated data $D^*$ is generated using the parameter vector $\theta^*$ for each particle. For the successful use of approximate Bayesian computation, choosing a distance function to compare simulated data $D^*$ to measurement data $D$ is crucial. A commonly used distance function is the normalised mean square error (NMSE). However, since we want to encourage sparsity in the parameter vector, we define the distance function (or loss) by adding an additional sparsity-promoting term to the NMSE. The loss $\rho(D,D^*)$ is defined as
% \begin{equation}
%     \rho(D,D^*) = \rho(D,D^*,\beta, \theta^*) = \sum_{j \in S} \frac{100}{m \sigma_j^2}\sum_{i=1}^{m} (\dot{x}^*_{j}(t_i)-\dot{x}_{j}(t_i))^2 + \beta ||\theta^*||_0
% \end{equation}
where $\sigma_{\bm D}^2$ is the variance of the measured data $\bm D$, $\beta$ is a sparsity-promoting factor, $\bm \theta^*$ is the parameter vector used to simulate data $\bm D^*$ and $||\bm \theta^*||_0$ is the L0 norm of $\bm \theta^*$. The generated particles are only accepted if the loss $\rho(\bm D, \bm D^*, \beta, \bm \theta^*)$ is within a predefined threshold $\epsilon_1$. 
The overall algorithm for generating the initial particles is shown in Algorithm \ref{alg:init_pop}.

\algrenewcommand\algorithmicrequire{\textbf{Input:}}
\algrenewcommand\algorithmicensure{\textbf{Output:}}

\begin{algorithm}[ht!] 
\caption{Initial population generator}
\label{alg:init_pop}
\begin{algorithmic}[1]
\Require{Data $\bm D$, dictionary of basis functions $\bm B$, total number of particles $N_S$, slap distribution $P_{\textrm{slab}}(\bm \theta)$; sparsity-promoting hyperparamters $\beta$, $\eta$ and $\bm \lambda = [\lambda_1, \lambda_2, ...., \lambda_{N}]$}
\Ensure{Initial population $\bm \Theta_1$, loss values $\bm E_1$ of the particles in the initial population}

%\Procedure{Generate initial population of particles}{}
    \State $p = 1$, $\textrm{iter} = 0$
    \While{$\textrm{iter} < N_S$}
        \State Sample parameters from prior: $\bm \theta^* \sim P_{\textrm{slab}}(\bm \theta) \in \mathbb{R}^N$
        \For{$i = $ $1$ to $N$}
            \State Sample random number $v \sim {\mathcal{U}}(0,1)$
            \If{$v > \eta$}
                \State $\theta^*_i = 0$ 
            \EndIf
            \If{$ |\theta^*_i|  < \lambda_i$}
                \State $ \theta^*_i = 0$ 
            \EndIf
        \EndFor
        \State Simulate data $\bm D^*$ using the parameters $\bm \theta^*$ \hfill \(\triangleright\) Eq. (\ref{eq:edps})
        \State Calculate the loss $\rho(\bm D, \bm D^*, \beta, \bm \theta^*)$ \hfill \(\triangleright\) Eq. (\ref{eq:rNMSE}) 
        \If{$\rho(\bm D, \bm D^*, \beta, \bm \theta^*)< \epsilon_1$}
            \State Save parameters $\bm \theta^*$ in $\bm \Theta_1$
            \State Save loss $\rho(\bm D, \bm D^*, \beta, \bm \theta^*)$ in $\bm E_1$
            \State $\textrm{iter} = \textrm{iter} +1$
        \EndIf
    \EndWhile

%\EndProcedure

\end{algorithmic}
\end{algorithm}

\subsection{Update scheme for s-ABC sampler}
Once the initial population is generated, the objective is to sequentially update the particles so as to minimize the cumulative loss defined in Eq. \eqref{eq:rNMSE}. We propose an update scheme that generates particles at each iteration while ensuring sparsity. The key idea here is to encourage exploitation; the best particles from the population in the previous iteration are utilized to generate better particles at each iteration while encouraging sparsity. We refer to the best particles used for the generation of the new particles as the active particles. 
Similar to initial population generation, we utilize the spike-and-slab prior introduced in Eq. \eqref{eq:ss}. However, a key difference here resides in the fact that the $P_{\text{slab}}\left( \bm \theta \right)$ is no longer uniformly distributed; instead, it is formulated by using the active particles,
\begin{equation}\label{eq:mog}
    P_{\text{slab}}\left( \bm \theta \right) = \sum_{i=1}^{K'} \phi_i \mathcal{N}(\mu_i, \Sigma_i).
\end{equation}
The optimal number of mixture components, $K'$, is decided by using the expectation-maximization algorithm in conjunction with Bayesian Information Criteria (BIC).
The procedure for generating $P_{\text{slab}}$ by using the active particles is shown in Algorithm \ref{alg:alg2}. 
\begin{algorithm}[ht!]
\caption{Construction of  Gaussian mixture model (GMM) for sampling}
\label{alg:alg2}
\begin{algorithmic}[1]
\Require{Current population of particles $\bm \Theta_p$, current loss values $\bm E_p$, next threshold $\epsilon_{p+1}$}
\Ensure{$(\phi_i$, $\mu_i$, $\Sigma_i)_{i=1}^{K'}$, $\bm A$, $\bm E_A$, $N_A$}
%\Procedure{Generate Gaussian mixture for sampling}{}
    \State Copy the particles with loss values less than $\epsilon_{p+1}$ to the set of active particles $\bm A = \{A_1, A_2, ..., A_{N_A} \}$%, and save the number of these particles ($N_A$)
    \State Save the loss values of the active particles to $\bm E_A$.
    \For{$K = 1$ to $K_{\textrm{max}}$}
        \State Randomly initialize $\bm \phi_K = \{\phi_1, ..., \phi_K \}$, $\bm \mu_K = \{\mu_1, ..., \mu_K \}$, $\bm \Sigma_K = \{\Sigma_1, ..., \Sigma_K \}$
        \While{not converged}
        \For{$i = 1$ to $N_A$}
        \For{$j = 1$ to $K$}
            \State{$r_{ij} = \dfrac{\phi_j \mathcal{N} (A_i | \mu_j, \Sigma_j)}{\sum_{j'=1}^K \phi_{j'} \mathcal{N} (A_i | \mu_{j'}, \Sigma_{j'})}$}

        \EndFor
        \EndFor

        \For{$j = 1$ to $K$}
            \State{$N_j = \sum_{i=1}^{N_A} r_{ij}$}
            \State{$\phi_j = \frac{N_j}{N}$}
            \State{$\mu_j = \frac{1}{N_j}\sum_{i=1}^{N_A} r_{ij} A_i$}
            \State{$\Sigma_j = \frac{1}{N_j}\sum_{i=1}^{N_A} r_{ij} (x_i - \mu_j) (x_i - \mu_j)^T $}
            
        \EndFor
        \EndWhile
    \State{Store parameters $(\bm \phi_K, \bm \mu_K, \bm \Sigma_K) = (\phi_i$, $\mu_i$, $\Sigma_i)_{i=1}^{K}$}
    \EndFor
    %\State Construct Gaussian mixtures $P_{\text{slab}} (\bm \theta_k) = \sum_{i=1}^K \phi_i \mathcal{N}(\mu_i, \Sigma_i)$ for $K = 1, 2, ..., K_{max}$ using the active particles $A$.
    \State Save the parameters $\{\bm \phi_{K'}, \bm \mu_{K'}, \bm \Sigma_{K'}\}$, where $K' \in \{1, 2, ..., K_{\textrm{max}}\}$ corresponds to the Gaussian mixture model with the lowest BIC.
    
%\EndProcedure

\end{algorithmic}
\end{algorithm}
The generated particles are 
% {are further filtered using sparsity-promoting thresholding as discussed before. An error threshold $\epsilon_{\text{new}}$ for selecting particles in the new population is defined, and the sparsity-modified sampled particles with loss $\rho(\bm D, \bm D^*, \beta, \bm \theta^*)$ less than $\epsilon_{\text{new}}$ along with the active particles from the previous population are selected to be part of the next population. } and 
further filtered by using Eq. \eqref{eq:rNMSE} as before. 
Algorithm \ref{alg:sampling} illustrates the generation of particles using thresholding and spike-and-slab prior with a mixture of Gaussians as slab distribution.
\begin{algorithm}[ht!]
    \caption{Sampling module to sample particles using thresholding and spike-and-slab prior with mixture of Gaussians as the slab distribution}\label{alg:sampling}
    \begin{algorithmic}[1]

\Require{ Parameters of the Gaussian mixture $(\phi_i$, $\mu_i$, $\Sigma_i)_{i=1}^{K'}$}%, Data $\bm D$ , current threshold $\epsilon_p$, number of active particles $N_A$, total number of particles in a population $N_S$}
\Ensure{Sampled particle $\theta^*$ %Sampled particles $\bm S$, loss values of the sampled particles $\bm E_S$}

%\Procedure{Sample particles from the Gaussian mixture}{}
    %\State $\textrm{iter} = 0$

    %\While{$\textrm{iter} < N_S-N_A$}
        \State Sample a particle from the Gaussian mixture : $\bm \theta^* \sim P_{\textrm{slab}} (\bm \theta) = \sum_{i=1}^{K'} \phi_i \mathcal{N}(\mu_i, \Sigma_i)$
        \State Sample random number $r \sim {\mathcal{U}}(0,1)$

        \If{$r>0.5$}
        \For{$i = $ $1$ to $N$}
            \State Sample random number $v \sim {\mathcal{U}}(0,1)$
            \If{$v > \eta$}
                \State $\theta^*_i = 0$ 
            \EndIf
            \If{$|\theta^*_i|  < \lambda_i$}
                \State $\theta^*_i = 0$ 
            \EndIf
        \EndFor
        \EndIf
        
        % \State Simulate data $\bm D^*$ using the parameters $\bm \theta^*$ \hfill \(\triangleright\) Eq. (\ref{eq:edps})
        % \State Calculate the loss $\rho(\bm D, \bm D^*, \beta, \bm \theta^*)$ \hfill \(\triangleright\) Eq. (\ref{eq:rNMSE}) 
        % \If{$\rho(\bm D, \bm D^*, \beta, \bm \theta^*)< \epsilon_p$}
        %     \State Save parameter $\bm \theta^*$ in $\bm S$
        %     \State Save loss $\rho(\bm D, \bm D^*, \beta, \bm \theta^*)$ in $\bm E_S$
        %     \State $\textrm{iter} = \textrm{iter} +1$
            
        % \EndIf

    %\EndWhile

%\EndProcedure
}
\end{algorithmic}
\end{algorithm}
Lastly, we adopt a similar strategy as the initial population generation by using Eq. \eqref{eq:rNMSE}.
{The proposed s-ABC sampler algorithm is presented in Algorithm \ref{alg:alg_sabc}}. 
% The overall procedure for the generation of new particles in a given iteration is shown in Algorithm \ref{alg:sampling}.
% \blue{The process of successively generating new populations (Algorithm \ref{alg:alg_sabc}) is continued till a stopping criterion is met. We define this criterion as being met when the difference between the error thresholds of two successively populations is less than a predefined defined error tolerance $\epsilon_{\text{tol}}$}

\algrenewcommand\algorithmicrequire{\textbf{Input:}}
\algrenewcommand\algorithmicensure{\textbf{Output:}}

\begin{algorithm}[ht!] 
\caption{Proposed s-ABC sampler}
\label{alg:alg_sabc}
\begin{algorithmic}[1]
%D = [\dot{x_2}(t_1), \dot{x_2}(t_2),..., \dot{x_2}(t_m)]
\Require{Data $\bm D$, dictionary of basis functions $\bm B$, initial population $\bm \Theta_1$, loss values of particles in the initial population $\bm E_1$, total number of particles $N_S$; sparsity-promoting hyperparamters $\beta$, $\eta$ and $\bm \lambda = [\lambda_1, \lambda_2, ..., \lambda_{N}]$; maximum number of components $K_{\textrm{max}}$ in the Gaussian mixture models, proportion of dropped particles $\alpha$, initial threshold $\epsilon_1$, stopping criterion}

%\Procedure{initial population generator}{$\bm D$, $N_S$, $\epsilon_1$}
%\State \textbf{Output: } $ \bm \Theta_1$, $\bm E_1$
%\EndProcedure

%\State Initial Population Generator($\bm D$, $N_S$, $\epsilon_1$) \(\rightarrow\) $ \bm \Theta_1$, $\bm E_1$ \hfill \(\triangleright\) Algorithm \ref{alg:init_pop} 

\State Sort $\bm E_1$ in descending order and get an integer $i_{f} = \textrm{floor}(\alpha N_S)$
\State Define the next threshold $\epsilon_2 = \bm E_{1,i_f}$ ($i_f$-th element of  $\bm E_1$)

%\Procedure{Construct Gaussian mixture for sampling}{$\bm \Theta_1$, $\bm E_1$, $\epsilon_2$}
%    \State \textbf{Output: } $(\phi_i$, $\mu_i$, $\Sigma_i)_{i=1}^{K'}$, $\bm A$, $\bm E_A$, $N_A$
%\EndProcedure

\State Construct GMM($\bm \Theta_1$, $\bm E_1$, $\epsilon_2$) \(\rightarrow\) $(\phi_i$, $\mu_i$, $\Sigma_i)_{i=1}^{K'}$, $\bm A$, $\bm E_A$, $N_A$ \hfill \(\triangleright\) Algorithm \ref{alg:alg2}

\While{stopping criteria is satisfied}
    \State Increment population $p = p+ 1$
    %\Procedure{Sample from Gaussian Mixture}{ $(\phi_i$, $\mu_i$, $\Sigma_i)_{i=1}^{K'}$, $\bm D$, $\epsilon_p$, $N_A$, $N_S$}
    %\State \textbf{Output: } $\bm S$, $\bm E_S$
    %\EndProcedure

    \State iter = 0, $\bm S$ = [ ],  $\bm E_s$ = [ ]
    \While{$\textrm{iter} < N_S-N_A$}
    
    \State Sample from GMM($(\phi_i$, $\mu_i$, $\Sigma_i)_{i=1}^{K'}$, $\bm D$)   \(\rightarrow\) $\bm \theta^*$ \hfill \(\triangleright\) Algorithm \ref{alg:sampling} 
    
    %\State $N_A$, $N_S$), $\bm E_S$ \hfill \(\triangleright\) Algorithm \ref{alg:sampling} 

    \State Simulate data $\bm D^*$ using the parameters $\bm \theta^*$ \hfill \(\triangleright\) Eq. (\ref{eq:edps})
    %\State Calculate the loss $\rho(\bm D, \bm D^*, \beta, \bm \theta^*)$ \hfill \(\triangleright\) Eq. (\ref{eq:rNMSE}) 
     \If{$\rho(\bm D, \bm D^*, \beta, \bm \theta^*)< \epsilon_p$} \hfill \(\triangleright\) Eq. (\ref{eq:rNMSE})
        \State Save parameter $\bm \theta^*$ in $\bm S$ and save loss $\rho(\bm D, \bm D^*, \beta, \bm \theta^*)$ in $\bm E_S$
         %\State Save loss $\rho(\bm D, \bm D^*, \beta, \bm \theta^*)$ in $\bm E_S$ 
        \State $\textrm{iter} = \textrm{iter} +1$
    \EndIf
    \EndWhile

    \State Define current population of particles: $\bm \Theta_p = \{\bm A, \bm S\}$
    \State Define loss values of the current population of particles: $\bm E_p = \{\bm E_A, \bm E_S\}$

    \State Sort $\bm E_p$ in descending order and get an integer $i_{f} = \textrm{floor}(\alpha N_S)$
    \State Define the next threshold $\epsilon_{p+1} = \bm E_{1,i_f}$ ($i_f$-th element of  $\bm E_1$)
    
    %\Procedure{Construct Gaussian mixture for sampling}{$\bm \Theta_p$, $\bm E_p$, $\epsilon_{p+1}$}
    %\State \textbf{Output: }  $(\phi_i$, $\mu_i$, $\Sigma_i)_{i=1}^{K'}$, $\bm A$, $\bm E_A$, $N_A$
    %\EndProcedure
    \State Construct GMM($\bm \Theta_p$, $\bm E_p$, $\epsilon_{p+1}$) \(\rightarrow\) $(\phi_i$, $\mu_i$, $\Sigma_i)_{i=1}^{K'}$, $\bm A$, $\bm E_A$, $N_A$ \hfill \(\triangleright\) Algorithm \ref{alg:alg2} 
    \State check stopping criterion 
    
\EndWhile
\end{algorithmic}
\end{algorithm}

\begin{figure}[ht!]
     \centering

         \includegraphics[scale = 0.5]{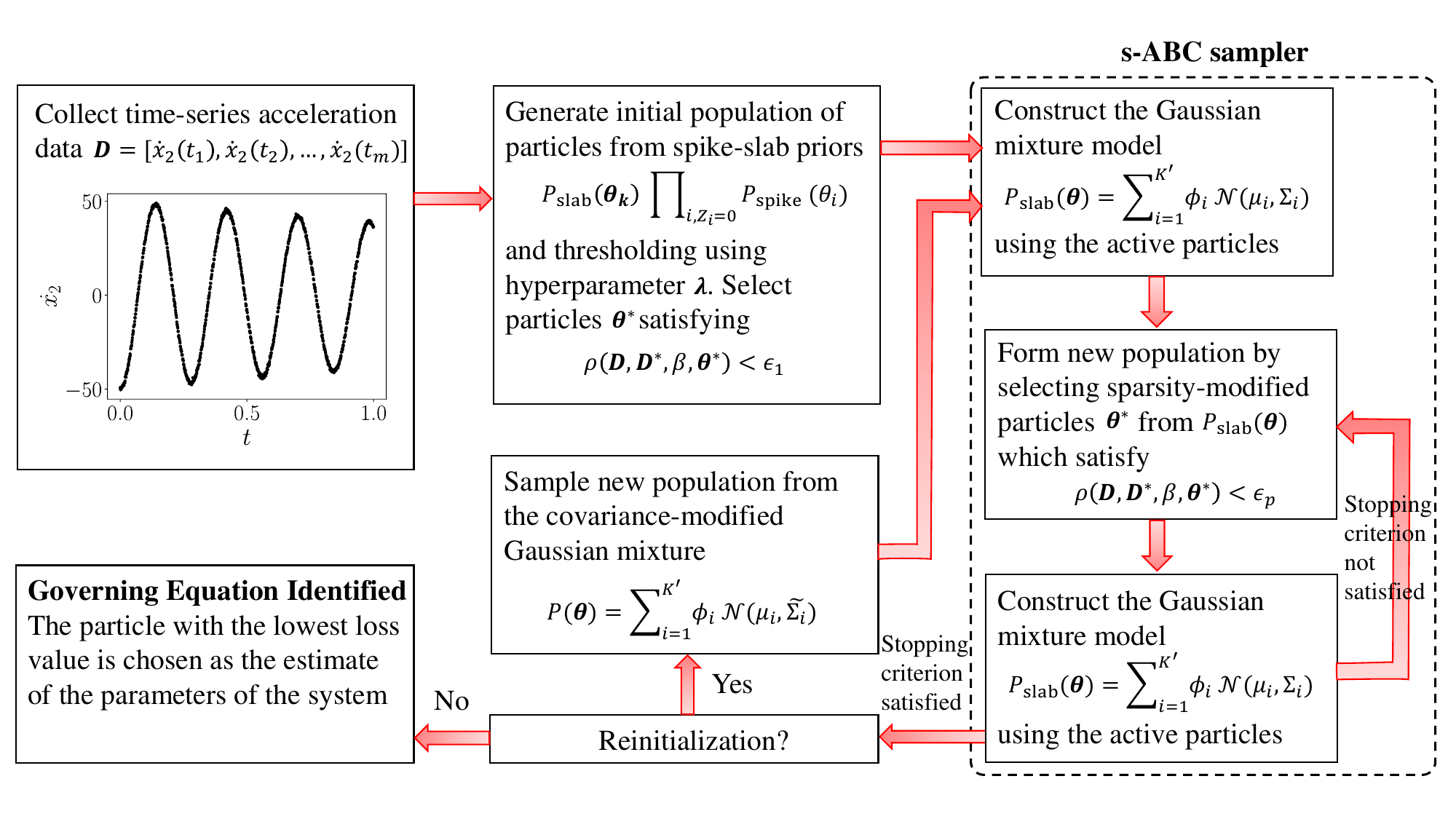}
         
         %\label{fig:fig_algo}
        \caption{Schematic illustration of the proposed s-ABC algorithm. First, acceleration data from the system is collected. Then, using predefined priors,  an initial population of particles is generated. Using the active particles from this population, a Gaussian mixture model for sampling is constructed. After this, a new population is formed using the active particles and newly sampled particles from this Gaussian mixture. Then, a Gaussian mixture using the active particles of the current population is constructed. The last two steps are repeated to generate new populations successively. This is done until the stopping criterion is satisfied. After this, if fine-tuning of the current parameters is necessary, then the process is reinitialization while encouraging exploration, and the process of constructing a Gaussian mixture model and sampling from it is repeated. If the estimated parameters are acceptable and reinitialization is not needed, then the parameters with the lowest loss value are selected to be the parameters of the identified model.}
        \label{fig:fig_algo}

\end{figure}

One of the key aspects of the proposed s-ABC algorithm is to find the correct balance between exploration and exploitation. The fact that slab density in Eq. \eqref{eq:mog} is formulated by using the active particles ensures exploitation. To ensure adequate exploration, we adopt a simple strategy wherein the covariance matrix of the Gaussian components in the mixture of Gaussians is first diagonalized, and the variance (diagonal elements of the covariance matrix) is enlarged,
\begin{equation}
    P\left( \bm \theta \right) = \sum_{i=1}^{K'} \phi_i \mathcal{N}(\mu_i, \tilde{\Sigma}_i),
\end{equation}
where
\begin{equation}
    \tilde {\Sigma}_i^{(kl)} = \left\{ \begin{array}{cl}
        \gamma  {\Sigma}_i^{(kl)} & \text{if }k=l, \\
         0 & \text{elsewhere.}
    \end{array} \right.
\end{equation}
This strategy encourages exploration by increasing the variance of the mixture components. Note that this strategy can also be used to develop an adaptive s-ABC algorithm; however, we here use this only for reinitialization. {The s-ABC sampler (Algorithm \ref{alg:alg_sabc}) is run multiple times with reinitialized particles to ensure a balance of exploration and exploitation.}  
On convergence, the particle with the lowest loss value is chosen as an estimation of the parameters of the identified model. The overall flow of the proposed s-ABC algorithm for equation discovery is shown in Fig. \ref{fig:fig_algo}.

\section{Numerical examples} \label{sec:sec4}
%In this section, we investigate the effectiveness of S-ABC in discovering the governing equations of various systems.
In this section, we present four numerical examples to illustrate the performance of the proposed algorithm. For all the examples, we generated synthetic data by solving the underlying governing equation by using the implicit Runge-Kutta method. The generated data is further corrupted by adding Gaussian noise to emulate a realistic scenario. We primarily examine two aspects: (a) the capability of the proposed model to capture the correct equation (i.e., correct terms from the library) and (b) the correct parameters associated with the candidate function. To evaluate (a), we calculate the inclusion probability using the final population. We also calculate the following error metric,
\begin{equation}\label{eq:fp}
    {\Delta_1}  = \frac{\tilde N_b - N_b}{N_b},
\end{equation}
where $\tilde N_b$ is the number of terms in the identified model (the best particle from the final population), and $N_b$ is the number of terms in the actual model. Note that for cases where the equation is identified correctly, {$\Delta_1=0$}. {$\Delta_1<0$} indicates that one or more terms are missing in the identified equation. On the other hand, {$\Delta_1>0$} indicates that additional terms are present in the identified equation. To evaluate (b), we utilize the mean squared relative error (MSRE) between the actual parameter and the identified parameters,
\begin{equation}
    { \Delta_2 = \frac{1}{N} \sum_{i=1}^N \left( \frac{(\hat{\theta}_i - \theta^{\textrm{true}}_i)}{\theta^{\textrm{true}}_i} \right)^2},
\end{equation}
% \begin{equation}
%      \Delta_2 = \frac{\sum_{i=1}^N \left(\hat \theta_i - \theta_i^{\text{true}} \right)^2}{\sum_{i=1}^N \left( \theta_i^{\text{true}} \right)^2}
% \end{equation}
{where $\hat{\theta}_i$ is the estimated parameter of the discovered model and $\theta^{\textrm{true}}_i$ is the true parameter of the actual model.}
We also illustrate the predictive performance of the proposed approach. To that end, we simulate the true and the identified equation by using a numerical integration scheme and compare the responses. 

\textit{Hyperparameter setup}: For all the numerical examples, we have considered $N_S = 400$, $\alpha = 0.05$, $\eta = 0.9$, $K_{\text{max}} = 5$. We define the stopping criterion as being reached when the difference between two successive error thresholds is less than a predefined error tolerance $\epsilon_{\text{tol}}$. Other problem-specific details are provided while the examples are described.

\subsection{Damped pendulum}
As the first example, we have considered a damped pendulum,
    \begin{equation}\label{eq:damped}
        \ddot{x} = T - \alpha \dot x - K \sin(x),
    \end{equation}
where $x$ is the angular position, $\dot x$ is the angular velocity, $T$ is the constant angular acceleration due to a constant driving force, $\alpha$ is the dissipation coefficient, and $K = g/l$, with $g$ being the acceleration due to gravity and $l$ being the length of the pendulum. For this example, we consider $T = 0.4$, $\alpha = 0.5$, $K=1$ with consistent units. Synthetic data is generated by simulating Eq. \eqref{eq:damped} for ten seconds with initial conditions $x(t=0) = 0, \dot x (t=0) = 0$. We consider that the sensor has a frequency of 30Hz; hence, we have 300 measurements equally spaced in the interval $[0,10]$. To emulate a realistic scenario, the simulated data generated is corrupted with 2\% Gaussian white noise. The objective here is to discover the underlying governing equation by using the noisy acceleration measurements.

To illustrate that the proposed s-ABC algorithm can discover the governing equation from acceleration-only measurements, we have created the library 
\begin{equation}
\bm B = \left[1, P^1\left(\bm z \right), P^2\left(\bm z \right), \ldots, P^5\left(\bm z \right), \sin \left(\bm z \right) \right] \in \mathbb R^{23},
\end{equation}
where $P^{\alpha}\left(\bm z \right)$ indicate all the terms present in $\left( x + \dot x\right)^{\alpha}$ and $\sin \left(\bm z \right)$ includes $\sin \left(x\right)$ and $\sin \left(\dot x\right)$. For initialization, we have considered 
\begin{equation}
    P_{\text{slab}} \left(\bm \theta\right) = \prod_j {P_{\text{slab}} \left( \theta _j \right)},
\end{equation}
where ${P_{\text{slab}} \left( \theta _j \right)} = \mathcal U \left(-1,1\right)$. The other hyperparameters of the s-ABC algorithms are set as follows: $\lambda = 0.2$, $\beta = 1$,  $\epsilon_1 = 10^5$, $\epsilon_{\text {tol}} = 0.005$. For reinitialization, we use $\gamma = 4$. Once reinitialized, we set $\epsilon_1 = 60$ and $\epsilon_{\text{tol}}=0.001$, while the remaining parameters are kept same.

\begin{figure}[ht!]
     \centering
         \includegraphics[scale = 0.65]{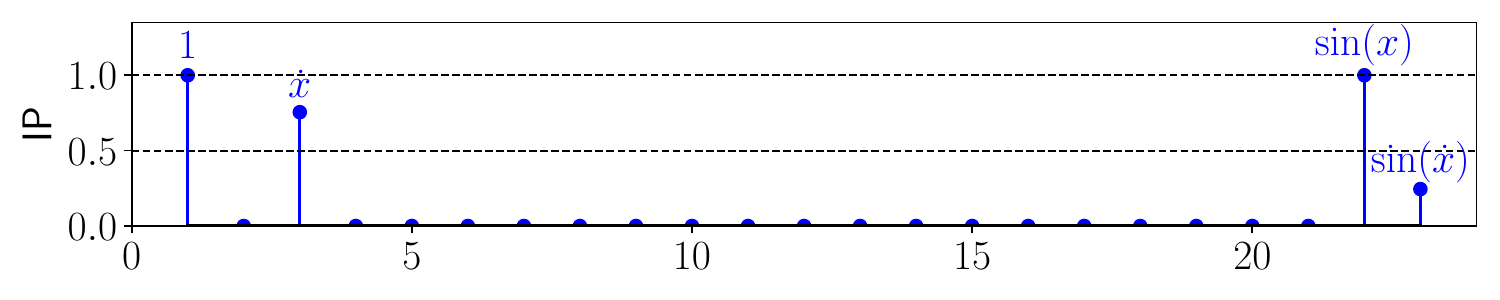}
        \caption{Inclusion probability (IP) of each candidate function in the final s-ABC population for the damped pendulum.}

      \label{fig:pend_acc_basis}
\end{figure}

To illustrate the performance of the proposed approach, we first concentrate on the ability of the proposed approach to identify the correct terms from the library $\bm B$. To that end, we calculate the inclusion probability (IP) of each term in $\bm B$ based on the final population. Fig. \ref{fig:pend_acc_basis} shows the IP of each of the components for this example. We observe that the proposed approach successfully identifies the correct terms with probability close to one. However, one additional term, $\sin (\dot x)$ has also been identified, although with relatively less probability. This can be attributed to the fact that $\dot x$ and $\sin (\dot x)$ are highly correlated for small $\dot x$. We select the best particle from the final population to arrive at the final model. For the final model, we obtain $\Delta_1 =0$. This indicates that the terms present in the final model matches exactly with those present in the actual model.  
% \blue{To arrive at the terms present in the identified equations, we select the parameter vector in the final population with the lowest loss. With this setup, we obtain $\Delta_1 =0$.}

\begin{figure}[ht!]     \centering
     \begin{subfigure}{0.49\textwidth}
         \centering
         \caption{}
         \includegraphics[scale = 0.495]{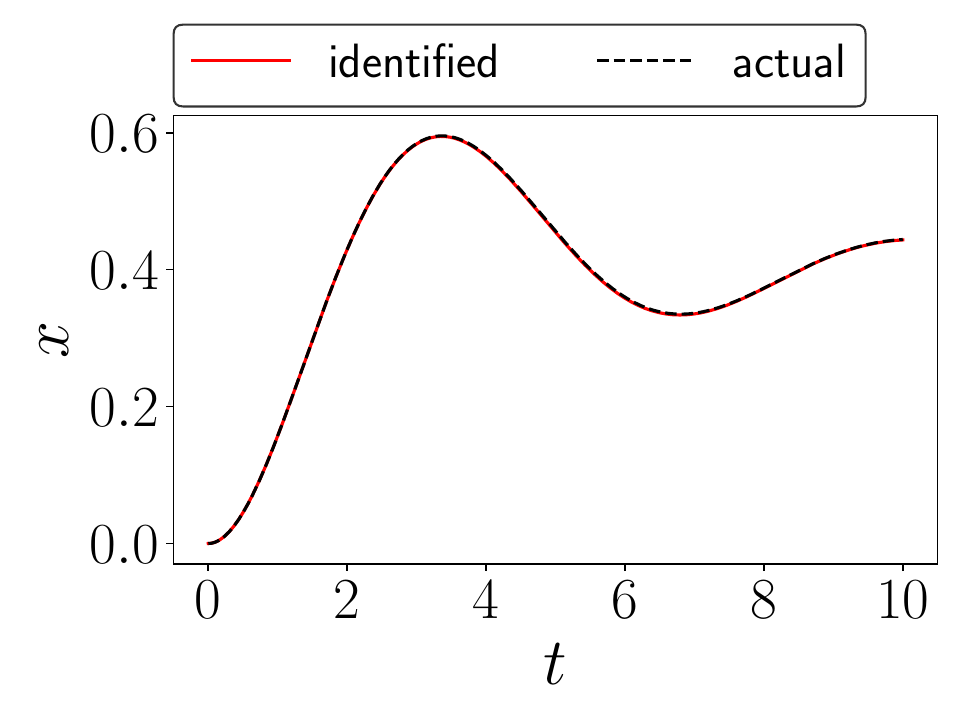}
         \label{fig:pend_acca}
     \end{subfigure}
     \hfill
     \begin{subfigure}{0.49\textwidth}
         \centering
         \caption{}
         \includegraphics[scale = 0.495]{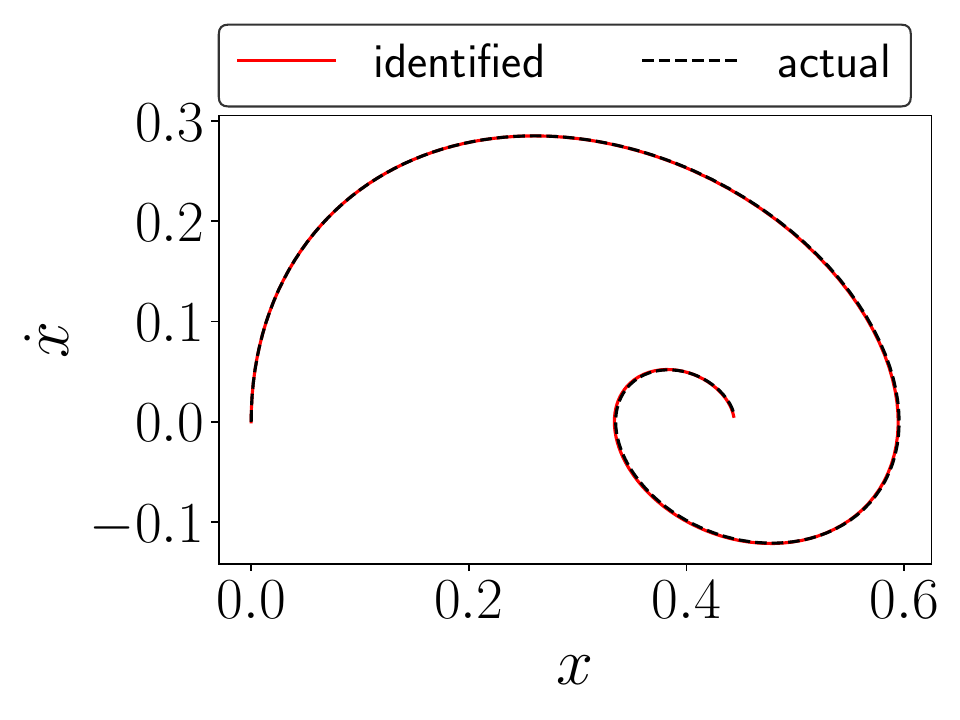}
         \label{fig:pend_accb}
     \end{subfigure}

    \hfill

        \caption{Simulated response of the actual model (black dashed line) and the discovered model (red line) for the damped pendulum.}
        \label{fig:pend_acc}
\end{figure}

Table \ref{tab:tabdamped1} shows the actual and the identified equation. We observe that the equation identified using the proposed approach is in close proximity to the actual equation (MSRE $\Delta_2 = 2.6875 \times 10^{-6}$). 
This reinforces our claim that the proposed s-ABC algorithm can identify governing equations from acceleration-only measurements. This is also evident from the fact that the response obtained using the identified equation matches exactly with those obtained using the actual equation (Fig. \ref{fig:pend_acc}).

\begin{table}[ht!]
    \centering
    \caption{The actual model, the discovered model, and the mean squared relative error (MSRE) between actual parameters and the predicted parameters for the damped pendulum.}
    \label{tab:tabdamped1}
    \begin{tabular}{l|l|l}
    \hline
        Actual Model & Discovered Model & $\Delta_2$ (MSRE)\\
        \hline \hline

         $\ddot{x}  =  0.4 - 0.5 \dot x -\sin (x) $ & $\ddot{x}  =  0.3999 - 0.4990 \dot x -1.002\sin(x)$ & $ 2.6875 \times 10^{-6}$ \\
         \hline
    \end{tabular}
\end{table}

\begin{comment}
\begin{figure}[h]
     \centering
 
         \includegraphics[scale = 0.495]{pendulum_acc.eps}
         \label{fig:fig4b}
        \caption{Measurement data (in black) and the simulated data of the predicted model (in red) for the damped driven pendulum.}
        \label{fig:pend_acc}
\end{figure}
\end{comment}

\subsection{Linear oscillator}
The next system that we examine is a linear oscillator given by
\begin{equation}\label{eq:linear}
        \ddot{x} =  -k x - c \dot{x},
\end{equation}
\begin{comment}
\begin{subequations}\label{eq:linear}
    \begin{equation}
        \dot{x} = y 
    \end{equation}
    \begin{equation}
        \dot{y} = -kx 0cy
    \end{equation}
\end{subequations}
\end{comment}
where $x$ is the displacement, $\dot{x}$ is the velocity, $k$ represents the stiffness (spring constant), and $c$ represents the damping coefficient. Similar to the previous example, the objective is to determine the governing equations using measurement data of the acceleration $\ddot{x}$. 

We have set the system parameters as $k=500$ and $c=0.5$, and have simulated this system for one second with the initial conditions $x (t=0) = 0.1, \dot{x}(t=0) = 0$. The system is solved in $[0,1]$ with time-step $\delta t = 10^{-3}$.
To further emulate the realistic conditions of sensor data collection, the saved values of $\ddot{x}$ have been corrupted with $2\%$ Gaussian noise.

% We have saved the simulated values of $\ddot{x}$ at $1000$ equally spaced time steps in the interval $[0, 1]$; these values mimic the measurements of a sensor with a frequency of 1000Hz. Then, to further emulate the realistic conditions of sensor data collection, the saved values of $\ddot{x}$ have been corrupted with $2\%$ Gaussian noise.

For this example, the s-ABC algorithm uses a dictionary of  21 potential candidate functions, 
\begin{equation}
    \bm B = [1, P^1(\bm z), P^2(\bm z), P^3(\bm z), P^4(\bm z), |\bm z|, \bm z|\bm z|] \in \mathbb{R}^{21},
\end{equation} 
where $P^{a}(\bm z)$ denotes all terms present in $(x + \dot{x})^a$. $|\bm z|$ denotes the terms $|x|$ and $|\dot{x}|$, and $\bm z|\bm z|$ denotes the terms $x|x|$, $x|\dot{x}|$, $\dot{x}|x|$ and $\dot{x}|\dot{x}|$. %The prior distributions $P_\text{slab} (\theta_i)$ for the parameters have been chosen as $\mathcal{U}[-1000,1000]$ for every parameter corresponding to a basis function with the term $x y^p |y|^q$ or $|x|  y^p |y|^q$, $\mathcal{U}[-10000,10000]$ for every parameter corresponding to a basis function with the term $x^2  y^p |y|^q$ or $x|x|  y^p |y|^q$, $\mathcal{U}[-100000,100000]$ for every parameter corresponding to a basis function with the term $x^3  y^p |y|^q$ and $\mathcal{U}[-1000000,1000000]$ for every parameter corresponding to a basis function with the term $x^4  y^p |y|^q$ (where $p = 0, 1, 2, 3, 4$ and $q = 0, 1$). The rest of the parameters have been sampled from the prior
For initialization, we have considered 
\begin{equation}
    P_{\text{slab}} \left(\bm \theta\right) = \prod_j {P_{\text{slab}} \left( \theta _j \right)}.
\end{equation}
We have considered that $P_\text{slab} (\theta_j)$ follow uniform distribution, $\mathcal U \left[-a,a\right]$. We have considered a partially informed prior such that
\begin{equation}\label{eq:prior_param}
    a = \left\{ \begin{array}{cl}
        100 \times 10^k & \text{for k-th order polynomial candidate function of }x \\
        1 & \text{elsewhere}
    \end{array} \right. ,
\end{equation}
where $k$ represents the order of polynomials. This prior encodes the information that high-order candidate functions have a wider search space. We further considered $\lambda_i = 0.2a$, where $a$ is obtained from Eq. \eqref{eq:prior_param}. The other hyperparameters have been set as follows: $\beta = 0.05$, $\epsilon_1 = 10^5$, $\epsilon_{tol} = 0.005$. Once the stopping criterion is met, we have reinitialized the population using $\gamma = 2$. After reinitialization, we have set $\beta = 0.05$, $\epsilon_1 = 20$, and $\epsilon_{tol} = 0.00001$, while the rest of the parameters retain the same value as before.  

% corresponding to the first order terms follow uniform distributon, $\mathcal{U}[-1000,1000]$, $\mathcal{U}[-10000,10000]$ for every parameter $(\theta_i)$ corresponding to a basis function with the factor $x^2$ or $x |x|$, $\mathcal{U}[-100000,100000]$ for every parameter $(\theta_i)$ corresponding to a basis function with the factor $x^3 $ and $\mathcal{U}[-1000000,1000000]$ for every parameter $(\theta_i)$ corresponding to a basis function with the factor $x^4$; else, $P_\text{slab} (\theta_i)$ is chosen as $\mathcal{U}[-1, 1 ]$. 

% The elements $\lambda_i$ of the hyperparameter vector $\bm \lambda \in \mathbb{R}^{N}$ are chosen to be $0.2$ times the upper bound of the prior uniform distribution $P_\text{slab} (\theta_i)$. The other hyperparameters have been set as follows: $\beta = 0.05$, $\epsilon_1 = 10^5$, $\epsilon_{tol} = 0.005$. Once the stopping criterion is met, we have reinitialized the population using $\gamma = 2$. After reinitialization, we have set $\beta = 0.5$, $\epsilon_1 = 20$, and $\epsilon_{tol} = 0.00001$, while the rest of the parameters remain the same as before.  

\begin{figure}[ht!]
     \centering
         \includegraphics[scale = 0.65]{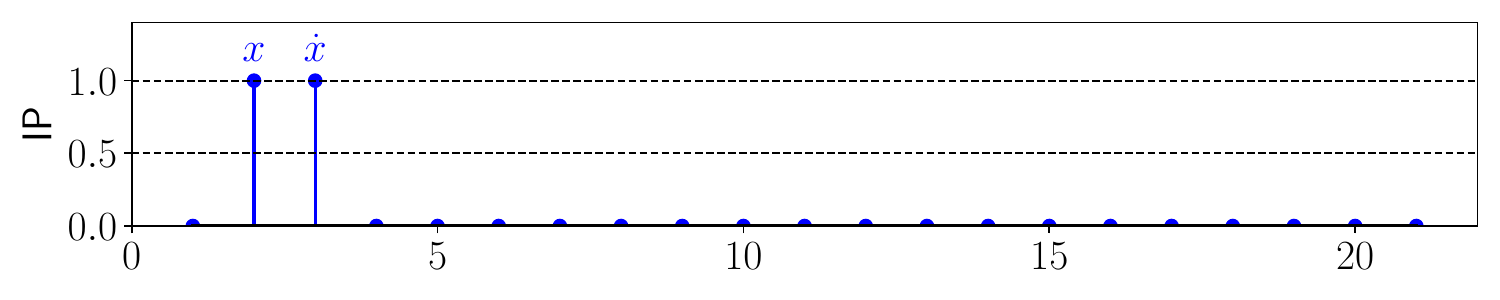}
        \caption{Inclusion probability (IP) of each basis function in the final s-ABC population for the linear spring-dashpot system.}

      \label{fig:fig_linbasis}
\end{figure}

Fig. \ref{fig:fig_linbasis} shows the IP of each candidate function in the final s-ABC population. It is observed that all particles in the final population contain the exact terms as in the actual equation, and hence, $\Delta_1 = 0$. This indicates the capability of the proposed approach to identify the correct candidates from the library $\bm B$. 
  
\begin{figure}[ht!]     \centering
     \begin{subfigure}{0.49\textwidth}
         \centering
         \caption{}
         \includegraphics[scale = 0.495]{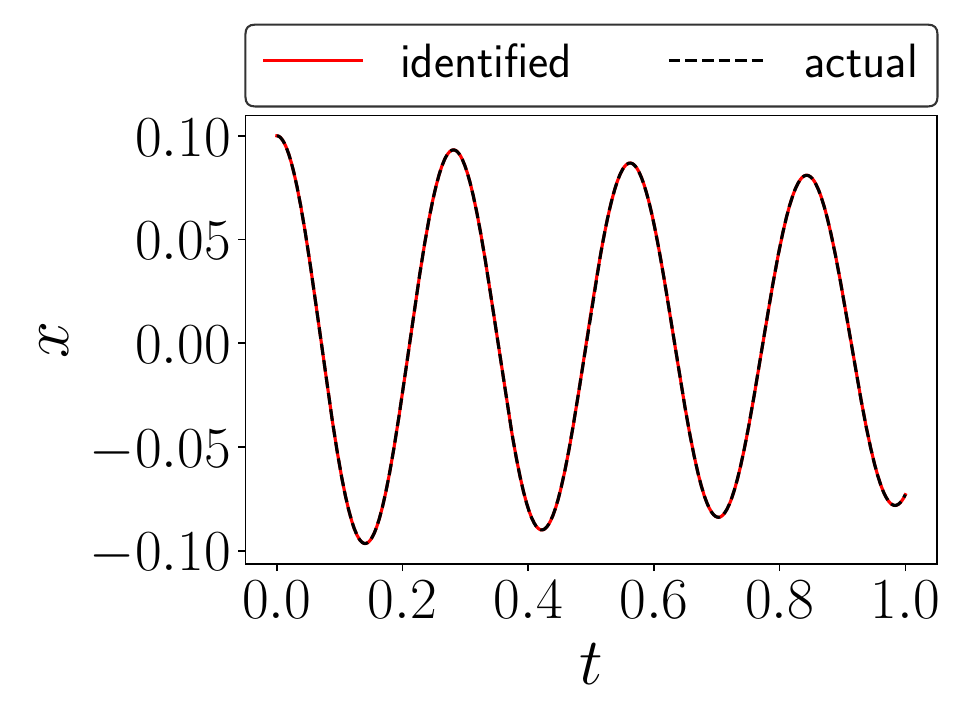}
         \label{fig:lineara}
     \end{subfigure}
     \hfill
     \begin{subfigure}{0.49\textwidth}
         \centering
         \caption{}
         \includegraphics[scale = 0.495]{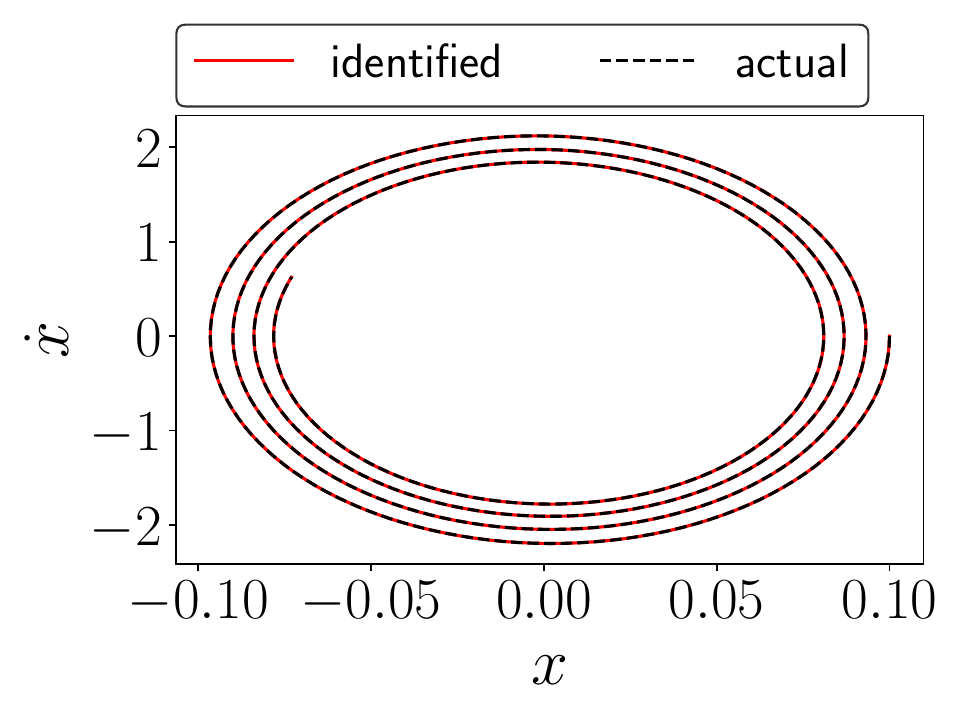}
         \label{fig:linearb}
     \end{subfigure}

    \hfill

        \caption{Simulated response of the actual model (black dashed line) and the discovered model (red line) for the linear spring-dashpot system.}
        \label{fig:linear}
\end{figure}

Table \ref{tab:linear} shows the actual equation, identified equation and the mean squared relative error (MSRE) between actual and identified parameters for this example. We observe that the identified parameters match closely with the actual parameters. This is also indicated by the fact that MSRE, $\Delta_2 = 1.8980 \times 10^{-6}$.  Fig. \ref{fig:linear} compares the response obtained from the actual model and the response from the identified equation. We observe that the two responses have a close match, further reinforcing the excellent performance of the proposed approach.
% The response obtained from the identified model closely matches that from the actual model, further indicating the success of s-ABC.

% The predicted parameters closely match the actual parameters. The error metric $\Delta_1 = 0$ as the exact terms from the actual model are present in the predicted model.

\begin{table}[ht!]
    \centering
    \caption{The actual model and the identified model, and the mean squared relative error (MSRE) between actual parameters and the predicted parameters for the linear spring-dashpot system.}
    \label{tab:linear}
    \begin{tabular}{l|l|l}
    \hline
        Actual model & Discovered model & $\Delta_2$ (MSRE)\\
        \hline \hline
          $\ddot{x}  =   -500x - 0.5\dot{x}$ &  $\ddot{x} =   -499.91x - 0.49903\dot{x}$ & $1.8980 \times 10 ^{-6}$ \\
         \hline 
 
    \end{tabular}
\end{table}

\begin{comment}
\begin{figure}[h]    
         \centering
         \includegraphics[scale = 0.495]{linear_acc.eps}
   
        \caption{Measurement data (in black) and the simulated data of the predicted model (in red) for the linear system.}
        \label{fig:linear}
\end{figure}
\end{comment}

\subsection{Duffing oscillator}
As the third example, we consider the Duffing oscillator. This is a nonlinear system with cubic nonlinearity in the displacement. The governing equation for the Duffing oscillator takes the following form,
\begin{equation}\label{eq:duffing}
        \ddot{x} =  -k x - c \dot{x} - k_3 x^3,
\end{equation}
\begin{comment}
\begin{subequations}\label{eq:linear}
    \begin{equation}
        \dot{x} = y 
    \end{equation}
    \begin{equation}
        \dot{y} = -k x - c y - k_3 x^3
    \end{equation}
\end{subequations}
\end{comment}
where $x$ is the displacement, $\dot{x}$ is the velocity, $k$ represents the stiffness, $c$ represents the damping, and $k_3$ is the coefficient associated with the nonlinear term. We have set the parameters $k=500$, $c=0.5$, $k_3 = 50000$. We have followed the same procedure and same setup as the previous example to generate synthetic data and have corrupted it with 2\% noise.
The goal again is to discover the above equation from noisy acceleration measurements.

% is to find an expression for $\ddot{x}$ using acceleration-only measurement data. 

% For this system,  To generate realistic synthetic acceleration data for this system, we have followed the exact same process as in the previous example.
%. $t=1$ with the initial condition $(x, y) = (0.1,0)$ using the model parameters in table \ref{tab:duffing} and have saved the values of $\dot{y}$ at $1000$ equally spaced time steps from $0$ to $1$. Then, the saved values of $\dot{y}$ have been corrupted with Gaussian noise having a standard deviation of 2\% of the standard deviation of the values $\dot{y}$; this forms the measurement data $\bm D$. 
We use the same dictionary of $21$ candidate functions as the previous example. The prior distribution $P_{\text{slab}}(\bm \theta)$ used for initialization and the hyperparameter vector $\bm \lambda \in \mathbb{R}^{N}$ have also been chosen to be identical to that in the case of the linear system. We have set the parameters $\beta = 0.05$, $\epsilon_1 = 10^5$, $\epsilon_{\text{tol}} = 0.005$. For reinitialization, we have used $\gamma = 2$ to create a new population. For the next iteration of the s-ABC sampler, we set the parameters $\beta = 0.5$, $\epsilon_1 = 20$, and $\epsilon_{tol} = 0.00001$, while the rest of the parameters remain the same as before. 

\begin{figure}[ht!]
     \centering
         \includegraphics[scale = 0.65]{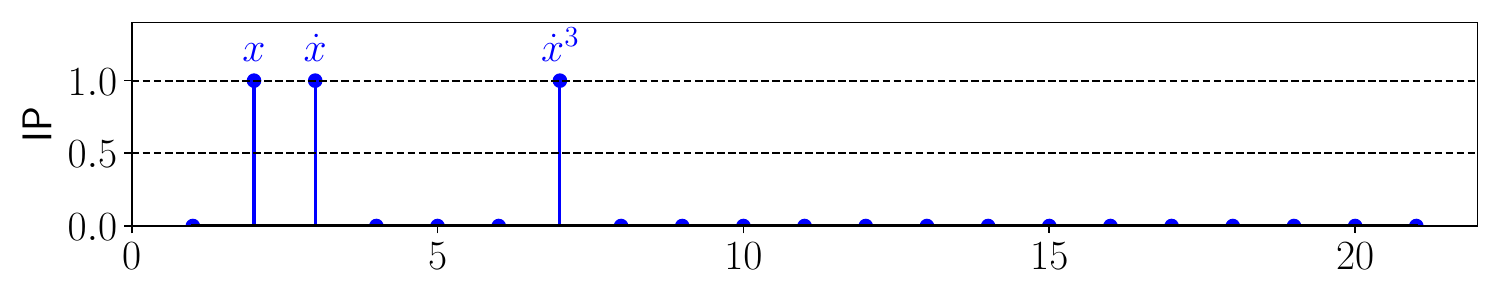}
        \caption{Inclusion probability (IP) of each candidate function in the final s-ABC population for the Duffing oscillator.}

      \label{fig:fig_duffbasis}
\end{figure}

Fig. \ref{fig:fig_duffbasis} shows the inclusion probability (IP) of each candidate function in the final s-ABC population. All particles in the final population contain the exact terms as in the actual equation. Therefore, $\Delta_1 =0$, indicating that the correct candidate functions have been correctly identified. 

\begin{figure}[ht!]     \centering
     \begin{subfigure}{0.49\textwidth}
         \centering
         \caption{}
         \includegraphics[scale = 0.495]{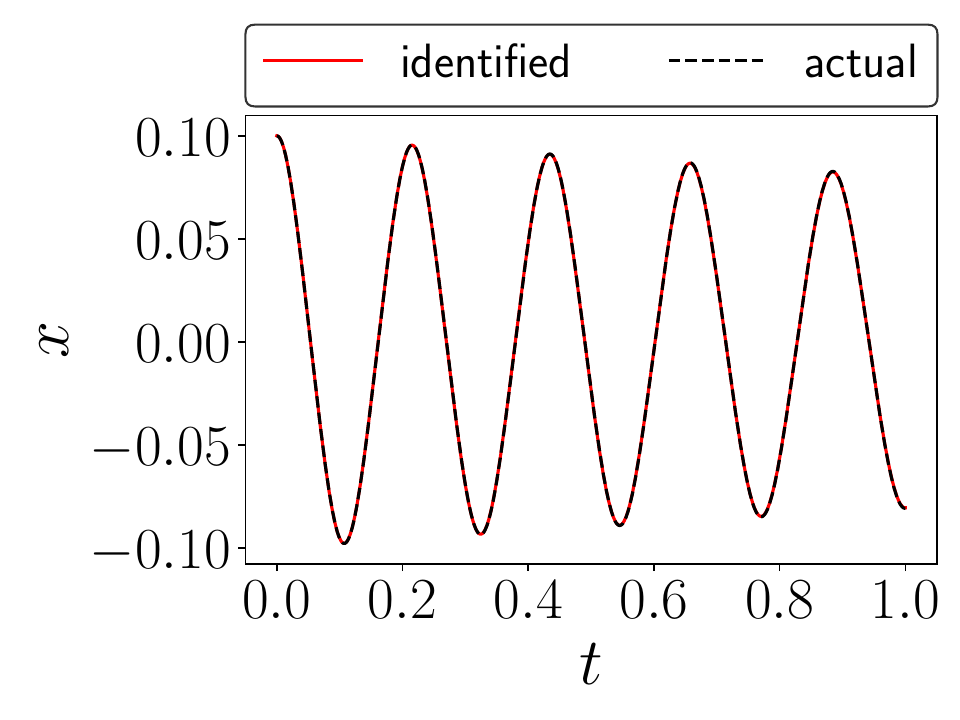}
         \label{fig:duffa}
     \end{subfigure}
     \hfill
     \begin{subfigure}{0.49\textwidth}
         \centering
         \caption{}
         \includegraphics[scale = 0.495]{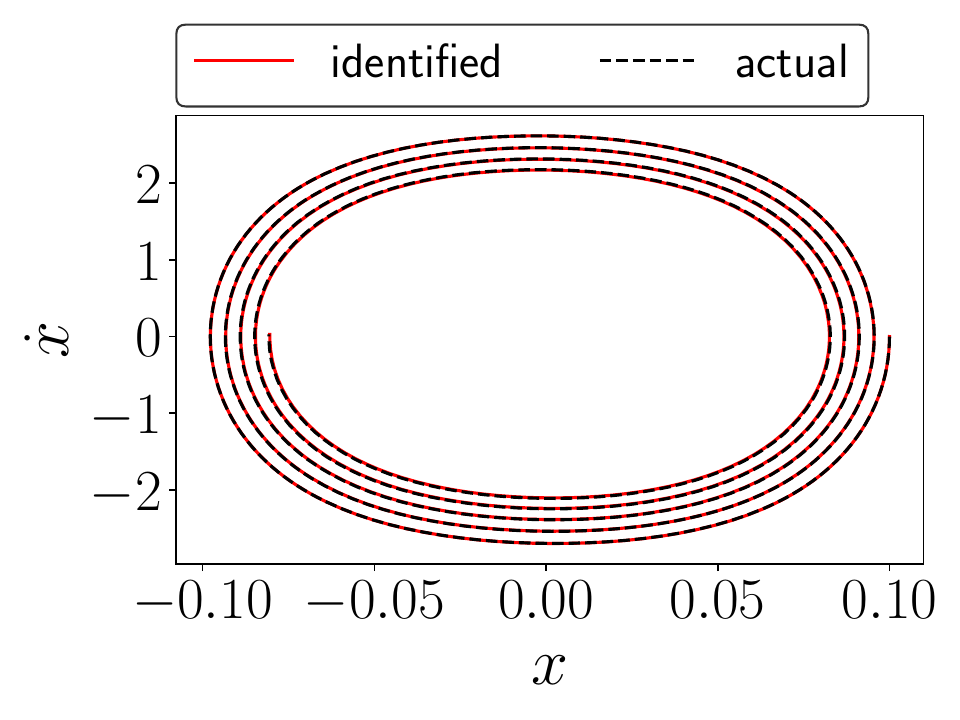}
         \label{fig:duffb}
     \end{subfigure}

    \hfill

        \caption{Simulated response of the actual model (black dashed line) and the identified model (red line) for the Duffing oscillator system.}
        \label{fig:figduff}
\end{figure}

Table \ref{tab:duffing} shows the actual equation, the predicted equation and the mean squared relative error (MSRE) between actual parameters and the predicted parameters model for the Duffing oscillator. The predicted parameters closely match the actual parameters, indicated by the fact that $\Delta_2 = 7.8674\times 10^{-5}$. Fig. \ref{fig:figduff} depicts the response obtained from the actual and discovered models. Both the responses closely match with each other.

\begin{table}[ht!]
    \centering
    \caption{The actual model, the identified model, and the mean squared relative error (MSRE) between actual parameters and the predicted parameters for the Duffing oscillator.}
    \label{tab:duffing}
    \begin{tabular}{l|l|l}
    \hline
         Actual model & Discovered model & $\Delta_2$ (MSRE) \\
        \hline \hline

        $\ddot{x} =   -500x - 0.5\dot{x} - 50000x^3$ & $\ddot{x}  =   -501.02x - 0.50761\dot{x} - 49977x^3$ & $7.8674 \times 10 ^{-5}$    \\
         \hline

    \end{tabular}
\end{table}

\begin{comment}
    
\begin{figure}[h]    
         \centering
         \includegraphics[scale = 0.495]{duffing_acc.eps}
   
        \caption{Measurement data (in black) and the simulated data of the predicted model (in red) for the Duffing system.}
        \label{fig:figduff}
\end{figure}

\end{comment}

\subsection{Oscillator with quadratic viscous damping}
As the final example, we consider an oscillator with quadratic viscous damping. This system is given by
\begin{equation}\label{eq:viscous}
    \ddot{x} = -k x - c \dot{x} - c_2 \dot{x} |\dot{x}|,
\end{equation}
where $x$ is the displacement, and $\dot{x}$ is the velocity. For this system, we have set the parameters $k=500$, $c=0.5$, $c_2 = 0.8$. To generate realistic synthetic acceleration measurement data for this system, we have followed the same process as in the case of the linear system and the Duffing system. The goal is to use s-ABC to determine the governing equation of this system using noisy acceleration measurements.

We have used the same dictionary of $21$ candidate functions as in the case of the previous two examples. Furthermore, the prior distribution $P_{\text{slab}}(\bm \theta)$ used for initialization and the hyperparameter vector $\bm \lambda \in \mathbb{R}^{N}$ are also identical to that in the case of the linear and Duffing systems. We have set the parameters $\beta = 0.05$, $\epsilon_1 = 10^5$, $\epsilon_{\text{tol}} = 0.001$. For reinitialization, we have used $\gamma = 2$ to create a new population. For the next iteration of the s-ABC sampler, we set the parameters $\beta = 0.005$, $\epsilon_1 = 50$, and $\epsilon_{tol} = 0.00001$, while the rest of the parameters remain the same as before.

%We have simulated this system for $t=1$ with the initial condition $(x, y) = (0.1,0)$ using the model parameters in table \ref{tab:viscous} and have saved the values of $\dot{y}$ at $1000$ equally spaced time steps from $0$ to $1$. Then, the saved values of $\dot{y}$ have been corrupted with Gaussian noise having a standard deviation of 2\% of the standard deviation of the values $\dot{y}$; this forms the measurement data $\bm D$. We have considered the same dictionary of $21$ library functions as in the case of the linear system and the Duffin system. The prior distributions $P_{\text{slab}}(\theta_i)$ for the parameters $\theta_i$ have been chosen to be identical to those in the case of the linear system and the Duffing system. 

\begin{figure}[ht!]
     \centering
         \includegraphics[scale = 0.65]{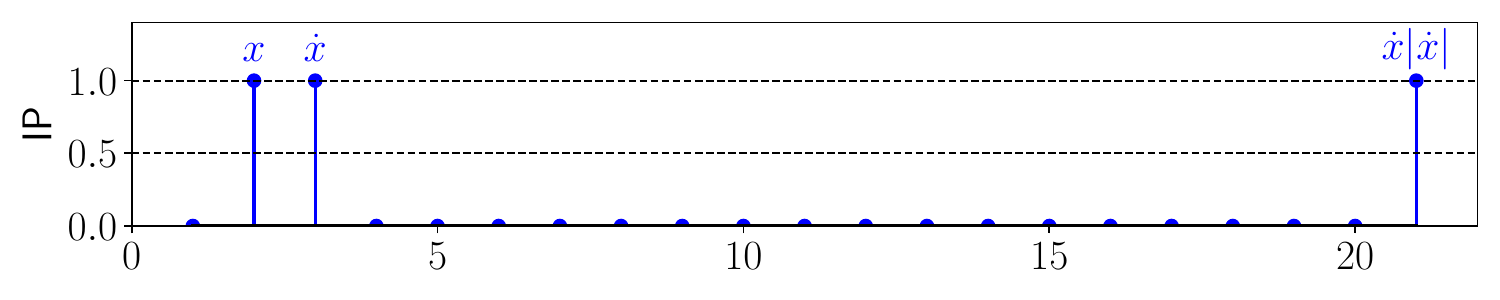}
        \caption{Inclusion probability (IP) of each candidate function in the final s-ABC population for the quadratic viscous damping oscillator problem.}

      \label{fig:fig_visbasis}
\end{figure}

Fig. \ref{fig:fig_visbasis} shows each candidate function's inclusion probability (IP) in the final s-ABC population. All particles in the final population only contain the terms present in the actual equation (i.e., $\Delta_1 = 0$). This indicates that the terms in the governing equations are correctly identified.

\begin{figure}[ht!]     \centering
     \begin{subfigure}{0.49\textwidth}
         \centering
         \caption{}
         \includegraphics[scale = 0.495]{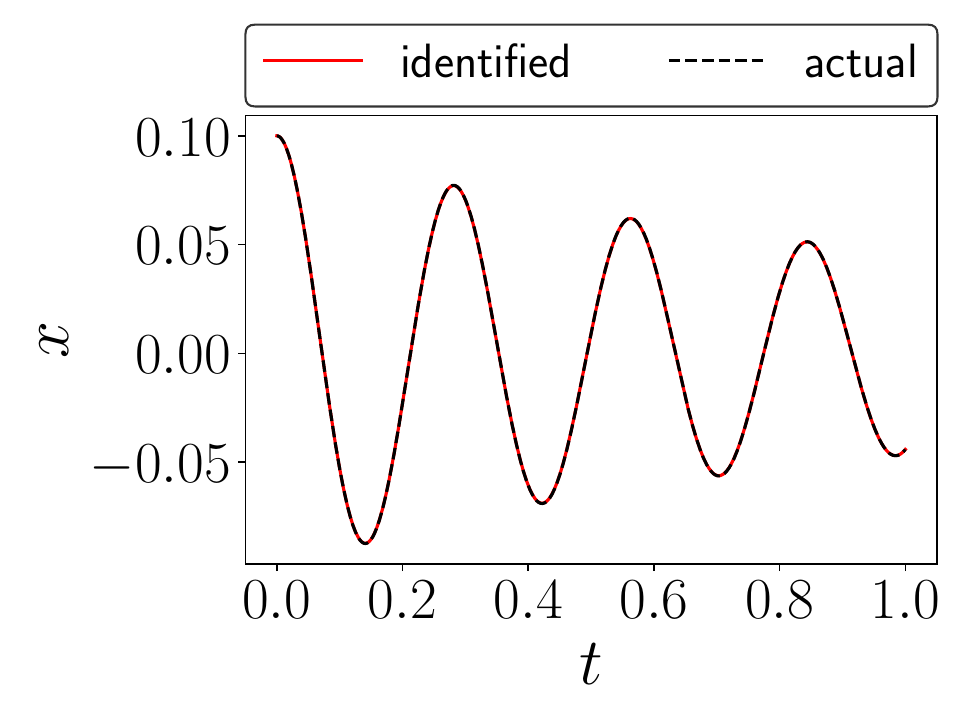}
         \label{fig:visca}
     \end{subfigure}
     \hfill
     \begin{subfigure}{0.49\textwidth}
         \centering
         \caption{}
         \includegraphics[scale = 0.495]{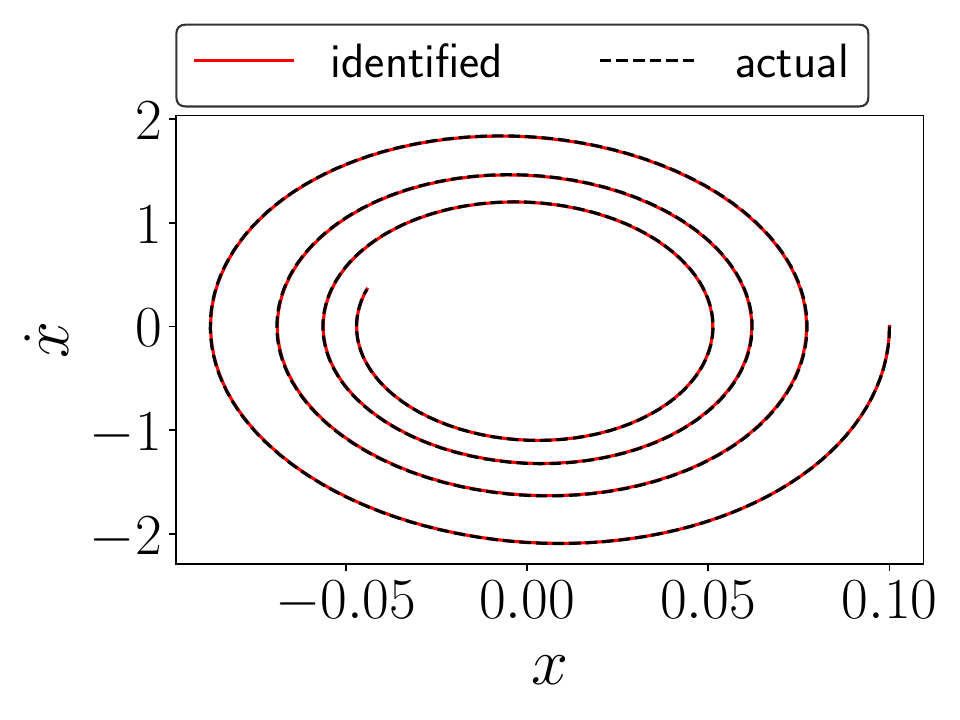}
         \label{fig:viscb}
     \end{subfigure}

    \hfill

        \caption{Simulated response of the actual model (black dashed line) and the identified model (red line) for the quadratic viscous damping system.}
        \label{fig:visc}
\end{figure}

Table \ref{tab:viscous} shows the actual equation, the discovered equation and the mean squared relative error (MSRE) between actual parameters and the predicted parameters model for the oscillator with viscous damping. The error metric $\Delta_1 = 0$ as the exact terms from the actual model are present in the identified model. Fig. \ref{fig:visc} depicts the simulated data from the actual and identified models. The identified model's response closely matches the actual model's response, further highlighting the potential of s-ABC in discovering the governing equations from noisy acceleration-only measurements.

\begin{table}[ht!]
    \centering
    \caption{The actual model, the identified model, and the mean squared relative error (MSRE) between actual parameters and the predicted parameters for the quadratic viscous damping oscillator problem.}
    \begin{tabular}{l|l|l}
    \hline
        Actual model & Discovered model & $\Delta_2$ (MSRE) \\
        \hline \hline
     
        $\ddot{x}  =   -500x - 0.5\dot{x} - 0.8 \dot{x}|\dot{x}|$ &  $\ddot{x} =   -500.10x - 0.46722\dot{x}^3 - 0.82760 \dot{x} |\dot{x}|$ &  $1.9628 \times 10^{-3}$ \\
         \hline 
         
    \end{tabular}
    \label{tab:viscous}
\end{table}

\begin{comment}
    
\begin{figure}[H]    
         \centering
         \includegraphics[scale = 0.495]{viscous_acc.eps}
   
        \caption{Measurement data (in black) and the simulated data of the predicted model (in red) for the quadratic viscous damping system.}
        \label{fig:visc}
\end{figure}
\end{comment}

\section{Discussion and Conclusion} \label{sec:sec5}
In this paper, we have proposed a novel Approximate Bayesian Computation (ABC) scheme to discover the governing equations of dynamical systems. Existing equation discovery methods require measurements of the state variables (position and velocity); this is a major bottleneck in structural dynamics and vibration where one often has access to acceleration, measured using accelerometers.
The proposed approach, referred to as sparse ABC (s-ABC), relaxes this requirement and allows the discovery of governing equations from acceleration. The proposed s-ABC algorithm uses a nested sampling approach in addition to multiple sparsity-promoting measures to progressively generate improved parameters that are closer to the system's actual parameters. The algorithm, when converged, provides the identified equations of the underlying system. Using four examples, which include a damped pendulum, a linear oscillator, a Duffing oscillator, and an oscillator with quadratic viscous damping, we have shown that s-ABC accurately discovers the governing equations from noisy measurements of acceleration.

% the proposed algorithm, termed s-ABC, uses acceleration-only data to uncover a system's governing equations.

% Unlike existing equation discovery methods, which require measurements of the state variables (position and velocity), the proposed algorithm, termed s-ABC, uses acceleration-only data to uncover a system's governing equations. The acceleration, velocity, and position can all be measured for a small system in a laboratory setup. However, for large structural systems, measuring all three variables is difficult. Acceleration is the most commonly measured quantity in structural testing, and hence, discovering the governing equations solely from acceleration data is the most realistic use case of equation discovery. Thus, the introduction of s-ABC could crucially contribute to the literature on equation discovery. The proposed s-ABC algorithm uses a nested sampling approach in addition to multiple sparsity-promoting measures to progressively generate improved parameters that are closer to the system's actual parameters. The algorithm, when converged, provides the identified equations of the underlying system. 

% Using four examples, which include a damped pendulum, a linear oscillator, a Duffing oscillator, and an oscillator with quadratic viscous damping, we have shown that s-ABC accurately discovers the governing equations from noisy measurements of acceleration. 

Despite the excellent performance of the proposed approach in all the examples, it is worthwhile to note that the proposed s-ABC has a large number of hyperparameters. 
For example, the hyperparameters $\eta$, $\beta$, and $\bm \lambda$ largely affect the identified model. Reinitialization also plays a crucial step in the equation discovery process.
While we have provided some practical values of the hyperparameters in this paper, further study is required on this aspect. Secondly, the proposed s-ABC algorithm falls under the broad umbrella of a library-based approach, and hence, for satisfactory performance, it is essential that the correct terms are included in the library. Also, highly correlated candidate functions present in the library may pose some challenges. This is observed in the first example where both $\dot x$ and $\sin (\dot x)$ have been identified for some particles in the final population. This can be attributed to the fact that for small $\dot x$, $\sin (\dot x) \approx \dot x$. Selective removal of highly correlated terms from the dictionary based on prior knowledge of the system can make equation discovery much easier.
Lastly, 
ABC algorithm necessitates multiple runs of the simulator. This can potentially make the algorithm computationally burdensome. Accelerating ABC algorithm is an open problem in the literature, and any success story along this line can potentially benefit the proposed approach as well.

\section*{Acknowledgements}
SC acknowledges the financial support received from the Ministry of Port and Shipping via letter no. ST-14011/74/MT (356529).

%Bibliography
%\bibliographystyle{unsrt}  
%\bibliography{references}  

\begin{thebibliography}{10}

\bibitem{brunton2016discovering}
Steven~L Brunton, Joshua~L Proctor, and J~Nathan Kutz.
\newblock Discovering governing equations from data by sparse identification of nonlinear dynamical systems.
\newblock {\em Proceedings of the national academy of sciences}, 113(15):3932--3937, 2016.

\bibitem{bongard2007automated}
Josh Bongard and Hod Lipson.
\newblock Automated reverse engineering of nonlinear dynamical systems.
\newblock {\em Proceedings of the National Academy of Sciences}, 104(24):9943--9948, 2007.

\bibitem{schmidt2009distilling}
Michael Schmidt and Hod Lipson.
\newblock Distilling free-form natural laws from experimental data.
\newblock {\em science}, 324(5923):81--85, 2009.

\bibitem{rudy2017data}
Samuel~H Rudy, Steven~L Brunton, Joshua~L Proctor, and J~Nathan Kutz.
\newblock Data-driven discovery of partial differential equations.
\newblock {\em Science advances}, 3(4):e1602614, 2017.

\bibitem{schaeffer2017learning}
Hayden Schaeffer.
\newblock Learning partial differential equations via data discovery and sparse optimization.
\newblock {\em Proceedings of the Royal Society A: Mathematical, Physical and Engineering Sciences}, 473(2197):20160446, 2017.

\bibitem{loiseau2018constrained}
Jean-Christophe Loiseau and Steven~L Brunton.
\newblock Constrained sparse galerkin regression.
\newblock {\em Journal of Fluid Mechanics}, 838:42--67, 2018.

\bibitem{loiseau2018sparse}
Jean-Christophe Loiseau, Bernd~R Noack, and Steven~L Brunton.
\newblock Sparse reduced-order modelling: sensor-based dynamics to full-state estimation.
\newblock {\em Journal of Fluid Mechanics}, 844:459--490, 2018.

\bibitem{mangan2016inferring}
Niall~M Mangan, Steven~L Brunton, Joshua~L Proctor, and J~Nathan Kutz.
\newblock Inferring biological networks by sparse identification of nonlinear dynamics.
\newblock {\em IEEE Transactions on Molecular, Biological and Multi-Scale Communications}, 2(1):52--63, 2016.

\bibitem{dam2017sparse}
Magnus Dam, Morten Br{\o}ns, Jens Juul~Rasmussen, Volker Naulin, and Jan~S Hesthaven.
\newblock Sparse identification of a predator-prey system from simulation data of a convection model.
\newblock {\em Physics of Plasmas}, 24(2), 2017.

\bibitem{sorokina2016sparse}
Mariia Sorokina, Stylianos Sygletos, and Sergei Turitsyn.
\newblock Sparse identification for nonlinear optical communication systems: Sino method.
\newblock {\em Optics express}, 24(26):30433--30443, 2016.

\bibitem{narasingam2018data}
Abhinav Narasingam and Joseph Sang-Il Kwon.
\newblock Data-driven identification of interpretable reduced-order models using sparse regression.
\newblock {\em Computers \& Chemical Engineering}, 119:101--111, 2018.

\bibitem{lai2019sparse}
Zhilu Lai and Satish Nagarajaiah.
\newblock Sparse structural system identification method for nonlinear dynamic systems with hysteresis/inelastic behavior.
\newblock {\em Mechanical Systems and Signal Processing}, 117:813--842, 2019.

\bibitem{kaiser2018sparse}
Eurika Kaiser, J~Nathan Kutz, and Steven~L Brunton.
\newblock Sparse identification of nonlinear dynamics for model predictive control in the low-data limit.
\newblock {\em Proceedings of the Royal Society A}, 474(2219):20180335, 2018.

\bibitem{kaheman2020sindy}
Kadierdan Kaheman, J~Nathan Kutz, and Steven~L Brunton.
\newblock Sindy-pi: a robust algorithm for parallel implicit sparse identification of nonlinear dynamics.
\newblock {\em Proceedings of the Royal Society A}, 476(2242):20200279, 2020.

\bibitem{schaeffer2017sparse}
Hayden Schaeffer and Scott~G McCalla.
\newblock Sparse model selection via integral terms.
\newblock {\em Physical Review E}, 96(2):023302, 2017.

\bibitem{champion2019data}
Kathleen Champion, Bethany Lusch, J~Nathan Kutz, and Steven~L Brunton.
\newblock Data-driven discovery of coordinates and governing equations.
\newblock {\em Proceedings of the National Academy of Sciences}, 116(45):22445--22451, 2019.

\bibitem{boninsegna2018sparse}
Lorenzo Boninsegna, Feliks N{\"u}ske, and Cecilia Clementi.
\newblock Sparse learning of stochastic dynamical equations.
\newblock {\em The Journal of chemical physics}, 148(24), 2018.

\bibitem{tripura2023bayesian}
Tapas Tripura and Souvik Chakraborty.
\newblock A bayesian framework for discovering interpretable lagrangian of dynamical systems from data.
\newblock {\em arXiv preprint arXiv:2310.06241}, 2023.

\bibitem{mangan2019model}
Niall~M Mangan, Travis Askham, Steven~L Brunton, J~Nathan Kutz, and Joshua~L Proctor.
\newblock Model selection for hybrid dynamical systems via sparse regression.
\newblock {\em Proceedings of the Royal Society A}, 475(2223):20180534, 2019.

\bibitem{rudy2019deep}
Samuel~H Rudy, J~Nathan Kutz, and Steven~L Brunton.
\newblock Deep learning of dynamics and signal-noise decomposition with time-stepping constraints.
\newblock {\em Journal of Computational Physics}, 396:483--506, 2019.

\bibitem{raissi2018multistep}
Maziar Raissi, Paris Perdikaris, and George~Em Karniadakis.
\newblock Multistep neural networks for data-driven discovery of nonlinear dynamical systems.
\newblock {\em arXiv preprint arXiv:1801.01236}, 2018.

\bibitem{raissi2018deep}
Maziar Raissi.
\newblock Deep hidden physics models: Deep learning of nonlinear partial differential equations.
\newblock {\em Journal of Machine Learning Research}, 19(25):1--24, 2018.

\bibitem{zhang2018robust}
Sheng Zhang and Guang Lin.
\newblock Robust data-driven discovery of governing physical laws with error bars.
\newblock {\em Proceedings of the Royal Society A: Mathematical, Physical and Engineering Sciences}, 474(2217):20180305, 2018.

\bibitem{nayek2021spike}
Rajdip Nayek, Ramon Fuentes, Keith Worden, and Elizabeth~J Cross.
\newblock On spike-and-slab priors for bayesian equation discovery of nonlinear dynamical systems via sparse linear regression.
\newblock {\em Mechanical Systems and Signal Processing}, 161:107986, 2021.

\bibitem{fuentes2021equation}
R~Fuentes, R~Nayek, P~Gardner, N~Dervilis, T~Rogers, K~Worden, and EJ~Cross.
\newblock Equation discovery for nonlinear dynamical systems: A bayesian viewpoint.
\newblock {\em Mechanical Systems and Signal Processing}, 154:107528, 2021.

\bibitem{more2023bayesian}
Kalpesh~Sanjay More, Tapas Tripura, Rajdip Nayek, and Souvik Chakraborty.
\newblock A bayesian framework for learning governing partial differential equation from data.
\newblock {\em Physica D: Nonlinear Phenomena}, 456:133927, 2023.

\bibitem{tripura2023sparse}
Tapas Tripura and Souvik Chakraborty.
\newblock A sparse bayesian framework for discovering interpretable nonlinear stochastic dynamical systems with gaussian white noise.
\newblock {\em Mechanical Systems and Signal Processing}, 187:109939, 2023.

\bibitem{north2022bayesian}
Joshua~S North, Christopher~K Wikle, and Erin~M Schliep.
\newblock A bayesian approach for data-driven dynamic equation discovery.
\newblock {\em Journal of Agricultural, Biological and Environmental Statistics}, 27(4):728--747, 2022.

\bibitem{mathpati2023mantra}
Yogesh~Chandrakant Mathpati, Kalpesh~Sanjay More, Tapas Tripura, Rajdip Nayek, and Souvik Chakraborty.
\newblock Mantra: A framework for model agnostic reliability analysis.
\newblock {\em Reliability Engineering \& System Safety}, 235:109233, 2023.

\bibitem{mathpati2024discovering}
Yogesh~Chandrakant Mathpati, Tapas Tripura, Rajdip Nayek, and Souvik Chakraborty.
\newblock Discovering stochastic partial differential equations from limited data using variational bayes inference.
\newblock {\em Computer Methods in Applied Mechanics and Engineering}, 418:116512, 2024.

\bibitem{marin2012approximate}
Jean-Michel Marin, Pierre Pudlo, Christian~P Robert, and Robin~J Ryder.
\newblock Approximate bayesian computational methods.
\newblock {\em Statistics and computing}, 22(6):1167--1180, 2012.

\bibitem{mitchell1988bayesian}
Toby~J Mitchell and John~J Beauchamp.
\newblock Bayesian variable selection in linear regression.
\newblock {\em Journal of the american statistical association}, 83(404):1023--1032, 1988.

\bibitem{csillery2010approximate}
Katalin Csill{\'e}ry, Michael~GB Blum, Oscar~E Gaggiotti, and Olivier Fran{\c{c}}ois.
\newblock Approximate bayesian computation (abc) in practice.
\newblock {\em Trends in ecology \& evolution}, 25(7):410--418, 2010.

\bibitem{beaumont2010approximate}
Mark~A Beaumont.
\newblock Approximate bayesian computation in evolution and ecology.
\newblock {\em Annual review of ecology, evolution, and systematics}, 41:379--406, 2010.

\bibitem{beaumont2002approximate}
Mark~A Beaumont, Wenyang Zhang, and David~J Balding.
\newblock Approximate bayesian computation in population genetics.
\newblock {\em Genetics}, 162(4):2025--2035, 2002.

\bibitem{akeret2015approximate}
Jo{\"e}l Akeret, Alexandre Refregier, Adam Amara, Sebastian Seehars, and Caspar Hasner.
\newblock Approximate bayesian computation for forward modeling in cosmology.
\newblock {\em Journal of Cosmology and Astroparticle Physics}, 2015(08):043, 2015.

\bibitem{drovandi2011estimation}
Christopher~C Drovandi and Anthony~N Pettitt.
\newblock Estimation of parameters for macroparasite population evolution using approximate bayesian computation.
\newblock {\em Biometrics}, 67(1):225--233, 2011.

\bibitem{technow2015integrating}
Frank Technow, Carlos~D Messina, L~Radu Totir, and Mark Cooper.
\newblock Integrating crop growth models with whole genome prediction through approximate bayesian computation.
\newblock {\em PloS one}, 10(6):e0130855, 2015.

\bibitem{liepe2014framework}
Juliane Liepe, Paul Kirk, Sarah Filippi, Tina Toni, Chris~P Barnes, and Michael~PH Stumpf.
\newblock A framework for parameter estimation and model selection from experimental data in systems biology using approximate bayesian computation.
\newblock {\em Nature protocols}, 9(2):439--456, 2014.

\bibitem{abdessalem2018model}
Anis~Ben Abdessalem, Nikolaos Dervilis, David Wagg, and Keith Worden.
\newblock Model selection and parameter estimation in structural dynamics using approximate bayesian computation.
\newblock {\em Mechanical Systems and Signal Processing}, 99:306--325, 2018.

\bibitem{toni2009approximate}
Tina Toni, David Welch, Natalja Strelkowa, Andreas Ipsen, and Michael~PH Stumpf.
\newblock Approximate bayesian computation scheme for parameter inference and model selection in dynamical systems.
\newblock {\em Journal of the Royal Society Interface}, 6(31):187--202, 2009.

\bibitem{abdessalem2019model}
A~Ben Abdessalem, N~Dervilis, D~Wagg, and K~Worden.
\newblock Model selection and parameter estimation of dynamical systems using a novel variant of approximate bayesian computation.
\newblock {\em Mechanical Systems and Signal Processing}, 122:364--386, 2019.

\bibitem{nayek2023identification}
R~Nayek, AB~Abdessalem, N~Dervilis, EJ~Cross, and K~Worden.
\newblock Identification of piecewise-linear mechanical oscillators via bayesian model selection and parameter estimation.
\newblock {\em Mechanical Systems and Signal Processing}, 196:110300, 2023.

\bibitem{pritchard1999population}
Jonathan~K Pritchard, Mark~T Seielstad, Anna Perez-Lezaun, and Marcus~W Feldman.
\newblock Population growth of human y chromosomes: a study of y chromosome microsatellites.
\newblock {\em Molecular biology and evolution}, 16(12):1791--1798, 1999.

\bibitem{marjoram2003markov}
Paul Marjoram, John Molitor, Vincent Plagnol, and Simon Tavar{\'e}.
\newblock Markov chain monte carlo without likelihoods.
\newblock {\em Proceedings of the National Academy of Sciences}, 100(26):15324--15328, 2003.

\bibitem{chiachio2014approximate}
Manuel Chiachio, James~L Beck, Juan Chiachio, and Guillermo Rus.
\newblock Approximate bayesian computation by subset simulation.
\newblock {\em SIAM Journal on Scientific Computing}, 36(3):A1339--A1358, 2014.

\end{thebibliography}

\end{document}